\documentclass[letterpaper]{article} 
\usepackage{aaai2026}  
\usepackage{times}  
\usepackage{helvet}  
\usepackage{courier}  
\usepackage[hyphens]{url}  
\usepackage{graphicx} 
\usepackage{natbib}  
\usepackage{caption} 
\usepackage{algorithm}
\usepackage{algorithmic}

\usepackage{newfloat}
\usepackage{listings}
\DeclareCaptionStyle{ruled}{labelfont=normalfont,labelsep=colon,strut=off} 
\floatstyle{ruled}
\newfloat{listing}{tb}{lst}{}
\floatname{listing}{Listing}
\usepackage{rotating}
\usepackage{enumerate}
\usepackage{booktabs}
\usepackage{multirow}
\usepackage{array}
\usepackage{subcaption}

\title{VP-Bench: A Comprehensive Benchmark \\for Visual Prompting in Multimodal Large Language Models}
\author{
    Mingjie Xu\textsuperscript{\rm 1}\equalcontrib,
    Jinpeng Chen\textsuperscript{\rm 2}\equalcontrib,
    Yuzhi Zhao\textsuperscript{\rm 2}\thanks{Corresponding Author.},
    Jason Chun Lok Li\textsuperscript{\rm 3},
    Yue Qiu\textsuperscript{\rm 4},
    Zekang Du\textsuperscript{\rm 4},
    Mengyang Wu\textsuperscript{\rm 5},
    Pingping Zhang\textsuperscript{\rm 2},
    Kun Li\textsuperscript{\rm 2},
    Hongzheng Yang\textsuperscript{\rm 5},
    Wenao Ma\textsuperscript{\rm 5},
    Jiaheng Wei\textsuperscript{\rm 1},
    Qinbin Li\textsuperscript{\rm 4},
    Kangcheng Liu\textsuperscript{\rm 6},
    Wenqiang Lei\textsuperscript{\rm 7}
}
\affiliations{
    \textsuperscript{\rm 1}The Hong Kong University of Science and Technology (Guangzhou)\\
    \textsuperscript{\rm 2}City University of Hong Kong\\
    \textsuperscript{\rm 3}The University of Hong Kong\\
    \textsuperscript{\rm 4}Huazhong University of Science and Technology\\
    \textsuperscript{\rm 5}The Chinese University of Hong Kong\\
    \textsuperscript{\rm 6}Hunan University\\
    \textsuperscript{\rm 7}Sichuan University\\
    mingjiexu@hkust-gz.edu.cn, jinpeng.chen@my.cityu.edu.hk, yzzhao2-c@my.cityu.edu.hk
}

\usepackage{bibentry}

\begin{document}

\maketitle

\begin{abstract}
Multimodal Large Language Models (MLLM) have enabled a wide range of advanced vision-language applications, including fine-grained object recognition and contextual understanding. When querying specific regions or objects in an image, human users naturally use ``Visual Prompts" (VP) like bounding boxes to provide reference. However, no existing benchmark systematically evaluates the ability of MLLMs to interpret such VPs. This gap raises uncertainty about whether current MLLMs can effectively recognize VPs, an intuitive prompting method for humans, and utilize them to solve problems. To address this limitation, we introduce VP-Bench, aiming to assess MLLMs’ capability in VP perception and utilization. VP-Bench employs a two-stage evaluation framework: Stage 1 examines models’ ability to perceive VPs in natural scenes, utilizing 30K visualized prompts spanning 8 shapes and 355 attribute combinations. Stage 2 investigates the impact of VPs on downstream tasks, measuring their effectiveness in real-world problem-solving scenarios. Using VP-Bench, we evaluate 28 MLLMs, including proprietary systems (e.g., GPT-4o) and open-source models (e.g., InternVL3 and Qwen2.5-VL). In addition, we provide a comprehensive analysis of factors affecting VP understanding, such as variations in VP attributes, question arrangement, and model scale. VP-Bench establishes a new reference framework for studying MLLMs’ ability to comprehend and resolve grounded referring questions.
\end{abstract}

\begin{links}
    \link{Datasets}{https://github.com/Endlinc/VP-Bench}
\end{links}

\section{Introduction}
\label{sec:intro}

The emergence of multimodal large language models (MLLM)~\cite{OpenAI2024GPT4V,OpenAI2024GPT4o,liu2023visual} has spurred research into their applications across diverse downstream tasks. For example, a user can verbally instruct the model to recognize furniture in indoor 3D scenes~\cite{zhou2024gala3d,zhang2024agent3d,zhou2024feature}, ground wild animals in their natural environment as depicted in an image~\cite{rasheed2024glamm,zhang2024llava}, use external knowledge to determine a piece of furniture's brand and price, and provide insights into the habits of those wild animals.
Beyond object detection via natural‑language queries, researchers have explored MLLMs’ ability to interpret user‑drawn annotations (e.g., freehand regions) and identify interactions involving target instances in context~\cite{cai2024vip,fu2024blink}. Consequently, visual prompts (VP), graphical cues such as bounding boxes and alphabet tags, have emerged to direct model attention, offering an intuitive alternative to verbal descriptions. However, MLLMs still underperform human annotators in grounded referring tasks with VPs. To quantify this gap, ViP-Bench~\cite{cai2024vip} was introduced, comprising 303 image–question pairs across eight VP types in practical scenarios (e.g., OCR, mathematical reasoning). 
Although ViP-Bench provides valuable insights into region reasoning, it does not assess, first, how perceptible different VP styles are to models. For example, while bounding boxes are ubiquitous, their low‑contrast or overly thin edges may be difficult for models to detect, limiting task accuracy. Moreover, it remains unclear whether adding distinctive corner markers to a bounding box VP would further enhance performance. Second, how variations in VP design affect downstream performance. For example, consider a lung region that is suspected to contain a malignant lesion. Is the application of an additional overlaying VP on this area more effective in directing the model's attention compared to specifying the region’s location within the verbal instruction, while still maintaining the contextual information?

\begin{figure*}
    \centering
    \includegraphics[width=\linewidth]{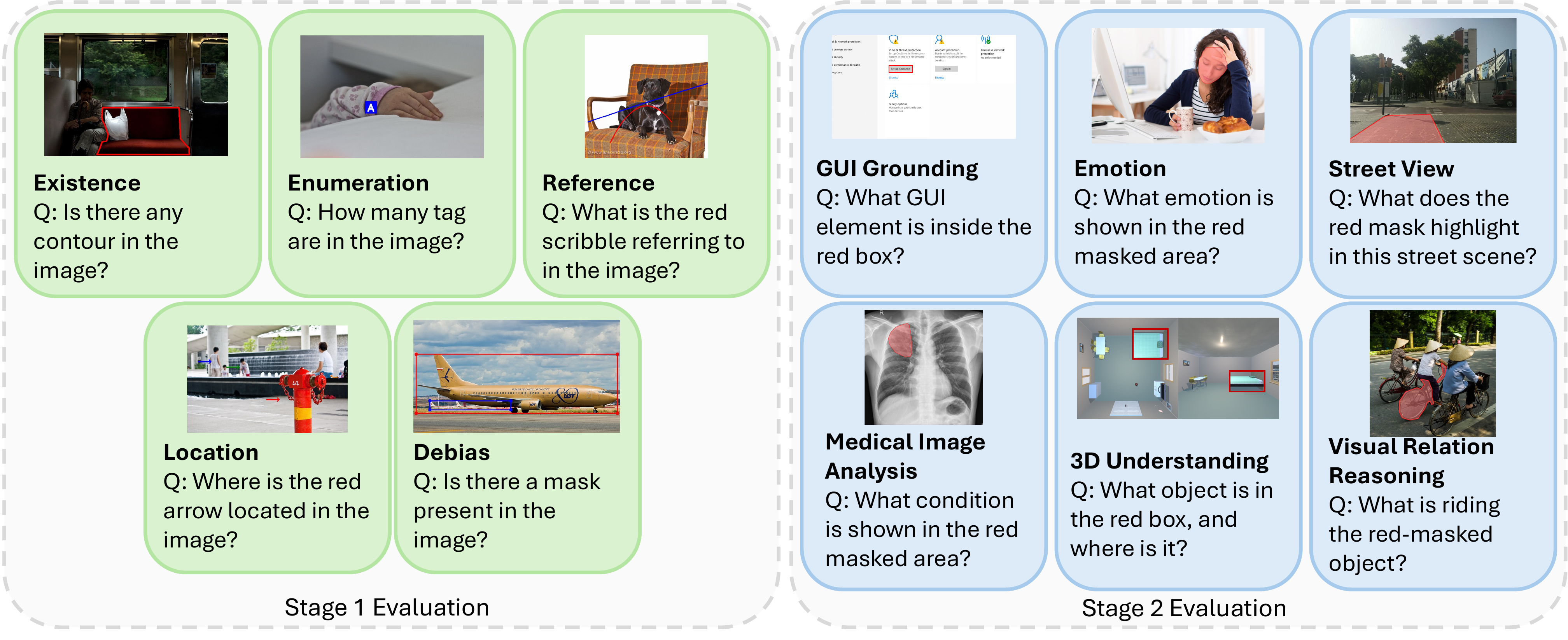}
    \caption{Overview of the VP-Bench Dataset. VP-Bench introduces a two-stage evaluation framework: (1) \textbf{Model Perception}, which assesses general VP recognition capabilities using 30K visualized VPs spanning five question types; and (2) \textbf{VP Effect on Downstream Tasks}, which evaluates the impact of visual prompts on various downstream applications. All questions follow a multiple-choice format, but the full list of options is not displayed due to space limitations.}
    \label{fig:dataset-overview}
\end{figure*}



To address these questions, we conduct a comprehensive review of prior studies~\cite{cai2024vip,fu2024blink,zhang2024llava,rasheed2024glamm,yang2023set} to categorize commonly employed VP shapes and their corresponding variants, which we subsequently define as attributes. We classify VP shapes into eight categories: tag, bounding box, arrow, mask, contour, oval, point, and scribble. Each category shares general attributes such as line width and color, and may also incorporate shape-specific attributes, including style or content format. A detailed overview of the VP shapes and their associated attributes is provided in the supplementary materials. In total, our proposed VP-Bench encompasses 355 unique VP attribute permutations. The benchmark consists of more than 30,000 distinct image–question pairs, constructed to evaluate core VP capabilities, including existence, enumeration, referencing, and spatial localization. This component constitutes Stage 1 of VP-Bench.
Furthermore, to examine the influence of VP on downstream task performance, we introduce six tasks that incorporate VP reasoning as a second stage, thereby reflecting the model’s ability to leverage VP reasoning in real-world problem-solving contexts. Additional examples from the benchmark are provided in Figure~\ref{fig:dataset-overview}, and a concise comparison between VP-Bench and other VP-related benchmarks is presented in Table~\ref{tab:benchmark_comparison}.

\begin{table}[t]
\centering
\small
\begin{tabular}{l c c c c}
\toprule
\textbf{Benchmark} 
  & \textbf{\# Images} 
  & \textbf{\# VP} 
  & \textbf{Domains} 
  & \textbf{Debias} \\
\midrule
SoV       
  & 119               
  & $\approx5$        
  & 1              
  & No               \\
ViP-Bench 
  & 303               
  & 8                 
  & 1             
  & No               \\
VP-Bench  
  & \textbf{34\,267}  
  & \textbf{355}      
  & 4
  & Yes              \\
\bottomrule
\end{tabular}
\caption{Comparison of three VP-related benchmarks. We list the number of images, number of VP attribute combinations, the image domains covered, and whether each dataset includes debias questions.}
\label{tab:benchmark_comparison}
\end{table}

We evaluate a range of popular MLLMs on our VP-Bench, including proprietary systems (e.g., GPT-4o) and open-source models (e.g., InternVL3 and Qwen2.5-VL). 
In Stage 1, we analyze the influence of factors such as instruction arrangement, model parameters, VP shapes, and attributes on model performance. In Stage 2, we examine the models’ capabilities across various VP-enabled downstream tasks, evaluating their ability to integrate visual cues with textual context, distinguish fine-grained object features, and leverage domain-specific knowledge. 
Moreover, we complement quantitative metrics with qualitative analyses to highlight each model’s interpretability and reliability in real-world scenarios. This comprehensive evaluation not only benchmarks current capabilities but also identifies potential areas for improvement, thereby informing future research in multimodal interaction and model interpretability.

To summarize, our contributions are as follows:
\begin{itemize}
    \item We introduce VP-Bench, a two-stage evaluation framework for assessing MLLMs. Stage 1 measures VP perception capability in natural scenes, while Stage 2 evaluates the ability to integrate VP understanding for practical problem-solving. Compared to existing benchmarks, our evaluation is significantly more comprehensive and includes over 100 times as many images.
    \item In stage 1, we examine the effects of 355 attribute combinations across eight VP shapes, covering a scale more than 40 times larger than previous studies. In Stage 2, we assess six VP-enabled downstream tasks, offering a comprehensive reference for real-world applications.
    \item Our results highlight the critical role of VP shape in model performance. Regular shapes (e.g., bounding boxes, ovals) are generally more efficient than irregular ones (e.g., masks, contours), even though the latter provide finer spatial details. Including VP shape descriptions in prompts further improves contextual understanding. Selecting VP shapes suited to the application scenario is more impactful than relying on the model’s preferred shape in a general scenario.
\end{itemize}

\section{Related Works}
\label{sec:relates}

\subsection{Multimodal Large Language Models}

Building on the recent success of neural language processing, particularly through LLM approaches, researchers have increasingly integrated visual understanding and reasoning to expand these models’ capabilities \cite{chen2022visualgpt,huang2023language,OpenAI2024GPT4V,li2023blip,liu2023visual}. Several studies have introduced visual instruction tuning \cite{liu2023visual} and specialized architectures \cite{li2023blip} for MLLMs, leading to significant advancements in image comprehension and common-sense reasoning. However, many existing MLLMs lack dedicated models or data designs for referring expressions or location-based referencing. Consequently, some researchers have adopted mask encoders to focus attention on specific regions \cite{Guo_2024_CVPR}, while others craft targeted text prompts to highlight region  \cite{wangall}.  
For instance, RegionGPT \cite{Guo_2024_CVPR} and LLaVA-Grounding \cite{zhang2024llava} employ additional mask encoders to improve location comprehension, though this adds computational overhead and demands retraining when introducing newer base models or additional VP shapes. Alternatively, approaches such as SoM \cite{yang2023set}, ViP-LLaVA \cite{cai2024vip}, and ControlMLLM \cite{wu2025controlmllm} demonstrate that sketching VPs directly on images can substantially enhance various downstream region-referring tasks. Yet, evaluations of these models’ handling of visually marked images have largely been qualitative, and comprehensive quantitative assessments across diverse region-referring tasks remain limited.

\subsection{MLLM Benchmarks}

Benchmarking MLLMs is crucial for exposing model limitations and guiding future development~\cite{yue2024mmmu,yu2024mm,mengmmiu,pmlr-v235-ying24a,liu2024mmbench,guan2024hallusionbench}. Although many existing benchmarks assess perception and reasoning, they largely emphasize image-level tasks. A few incorporate referring expression questions~\cite{wei2025large,zhang2024llava,li2025migician}, yet often neglect the role of VPs in visual understanding. For instance, RefCOCO~\cite{kazemzadeh2014referitgame} evaluates referring expression capabilities but lacks VP-oriented image design, while HC-RefLoCo~\cite{wei2025large} extends expression length without addressing the contribution of VPs to regional comprehension in MLLMs.
Recently, researchers have worked on developing VP-oriented benchmarks, such as the SoV \cite{zhang2024visual} validation dataset and ViP-Bench \cite{cai2024vip}, to provide a more thorough evaluation of VP cognition. However, these benchmarks still fall short when it comes to assessing the impact of VP on MLLM awareness and its effect on downstream tasks. For instance, while ViP-Bench offers a relatively comprehensive evaluation of VPs, it does not fully address the variations in VP shapes, attributes, and their effectiveness. These factors are crucial to understanding how different VPs influence MLLMs’ visual comprehension and their potential to improve downstream task performance. 

In contrast, our VP-Bench combines a broader range of VP shapes, attributes, and colors with a thorough examination of how VPs affect MLLM performance on various tasks. This comprehensive approach enables a more detailed assessment of model capabilities, providing insights into how VPs can be optimized for better task outcomes.

\section{VP-Bench}
\label{sec:method}

\begin{figure*}[t]
  \centering
  \includegraphics[width=\linewidth]{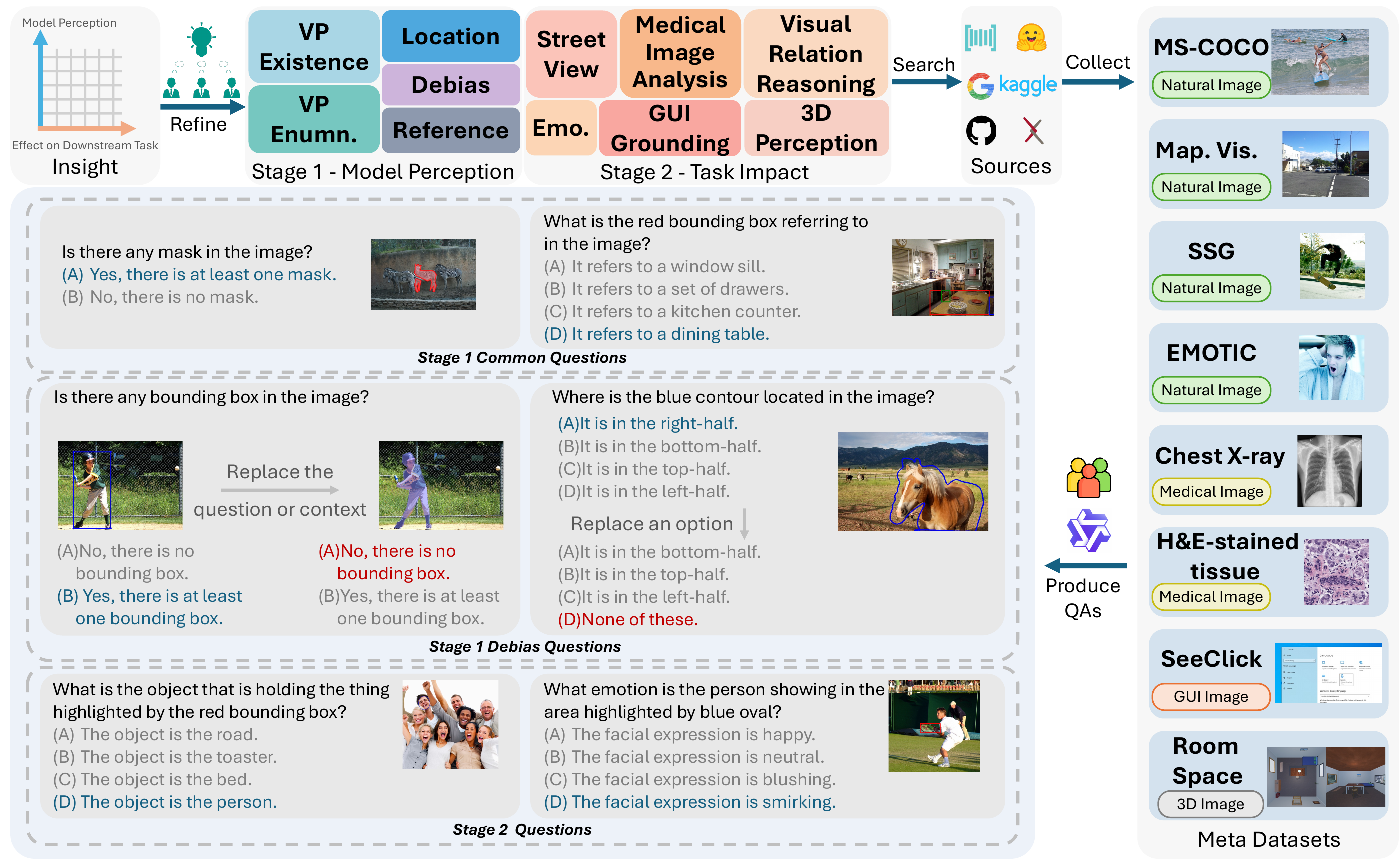}
  \caption{An illustration of our pipeline for data collection.
  Stage 1 is used to determine the general capabilities of MLLMs in recognizing VPs, while Stage 2 clarifies the impact of using VPs on downstream tasks.}
  \label{fig:data-collection}
\end{figure*}

\begin{table}[t]
  \centering
  \small
  \begin{tabular}{ll}
    \toprule
    \multicolumn{2}{c}{\textbf{Key Statistics}} \\ 
    \midrule
    \textbf{Statistic} & \textbf{Number} \\
    \midrule
    Total samples  & 38,932 \\
    Total images   & 34,267 \\
    \midrule
    \multicolumn{2}{c}{\textbf{VP Properties}} \\
    \midrule
    Shapes         & 8 \\
    Attributes     & 78 \\
    Colors         & 5 \\
    \midrule
    \multicolumn{2}{c}{\textbf{Debias Question}} \\
    \midrule
    Without Visualized VP          & 4.5\% \\
    Incorrect Model VP Instruction & 8\% \\
    \bottomrule
  \end{tabular}
  \caption{Key statistics for the dataset. This table provides an overview of the total number of samples, images, and tasks, as well as the debias question amount over the total benchmark.}
  \label{tab:key_statistics}
\end{table}

\subsection{Overview}
\label{sec:bench-overview}
VP-Bench is designed to assess the perception of VPs by MLLMs and to evaluate the impact of VPs on downstream task performance. Specifically, VP-Bench comprises 34,267 images and 38,932 multiple-choice questions. In constructing VP-Bench, we meticulously categorized VP shapes and attributes and developed a two-stage evaluation protocol. The Stage 1 data curation process assesses VP perception, while the Stage 2 data curation process examines the influence of VPs on models’ regional perception in downstream tasks.

Compared to existing VP-related benchmarks, VP-Bench introduces several key improvements (see Table \ref{tab:benchmark_comparison}). First, by incorporating a substantially larger set of test samples, VP-Bench covers an extensive range of VP attribute combinations and diverse grounded referring expression tasks. In Stage 1, it evaluates MLLMs’ perception across all 355 VP attribute combinations, exceeding the scope of other benchmarks by more than 40 times. This diversity compels models to develop a deeper understanding of visual information across various VP types. Second, in Stage 2, we determine the optimal VP combinations for each model and employ them to evaluate six downstream tasks within grounded referring scenarios, a critical gap that previous VP-focused frameworks have yet to fully address. Additionally, we introduce debias questions, where either the images lack the VP type mentioned in the question or all substantively descriptive answer choices are incorrect (see Figure \ref{fig:data-collection}) to mitigate unreliable conclusions arising from MLLMs’ hallucinations regarding VP presence. Detailed statistics of VP-Bench are provided in Table \ref{tab:key_statistics}.

Leveraging the rich data available in VP-Bench, our analytical framework enables a comprehensive evaluation of both VP perception and VP-based referring tasks. The top-down hierarchical structure of VP attributes allows for comparative analyses of models’ perception capabilities across various attributes. Moreover, the diverse downstream referring expression tasks facilitate performance estimation in practical scenarios, helping to delineate in-domain from out-of-domain tasks. Finally, the evaluation samples can be used to assess task learning difficulty, offering valuable insights for optimizing model training and dataset design. A detailed data curation breakdown is provided in the complementary material.

\begin{table}[h]
\centering
\small
    \begin{tabular}{p{2cm}p{5cm}}
        \toprule
        \textbf{VP Shape} & \textbf{Context} \\
        \midrule
        Tag & A tag is a small label located at the center of a target, displaying a number or letter. It may be in red, blue, green, or black, and can be circular or square in shape. \\
        Bounding Box & A bounding box is a rectangular frame that marks a target, which may be in red, blue, green, or black. \\
        Arrow & An arrow is a symbol that points to a target, which may be in red, blue, green, or black. \\
        Mask & A mask is a filled area used to indicate a target region, which may be in red, blue, green, or black. \\
        Contour & A contour is the outline of a target, which may be in red, blue, green, or black. It can be drawn precisely along the outline or may resemble a loosely hand-drawn line. \\
        Oval & An oval is an elliptical shape that encircles a target, which may be in red, blue, green, or black. \\
        Point & A point is a square or circular dot that represents a target, located at the center of the marked target, which may be in red, blue, green, or black. \\
        Scribble & A scribble is a random hand-drawn line that indicates a target, which may be in red, blue, green, or black. \\
        \bottomrule
    \end{tabular}
\caption{Definitions of the eight visual prompt (VP) shapes used in our benchmark. Each shape provides a distinct spatial cue (e.g., location, extent, or outline of the target) that is embedded into the model instructions.}
\label{tab:visual_prompt_shapes}
\end{table}

\subsection{Visual Prompt Description}
The VP description is designed to explicitly verbalize the spatial cues conveyed by visual prompts, rather than relying on the model to infer their meaning solely from the image. As summarized in Table~\ref{tab:visual_prompt_shapes}, each VP shape encodes a distinct type of spatial signal (e.g., region extent, outline). However, without a corresponding textual description, these cues remain implicit and may be underutilized by current MLLMs.
To address this gap, we propose a unified VP description scheme that converts each visual prompt into a short, structured phrase appended to the instruction. For example, a Bounding Box is described as ``the red box outlining the target region''. This mapping makes the semantics of each VP shape explicit in language.
This design offers two key benefits. First, it reduces ambiguity by explicitly stating whether a mark denotes a point or region. Second, it improves vision–language alignment by mirroring the same spatial cue in both the image and the text, guiding the model to the correct region.
Empirically, we observe consistent gains from adding VP descriptions, especially for complex shapes, showing that explicit textual VP semantics are key to fully leveraging visual prompts in MLLMs.

\subsection{Stage 1 Data Curation Process} \label{sec:stage1-curation}

VPs play a pivotal role in guiding model attention and addressing complex problems. In Stage 1 evaluation, we treat VPs as visual cues defined by their intrinsic shapes, attributes, and colors, enabling more targeted information retrieval. We design the VP attributes according to a top-down hierarchical structure. Initially, we categorize VPs by shape, such as tag, bounding box, arrow, mask, contour, oval, point, and scribble, and further subdivide them by finer-grained attributes and colors. The specific VP shapes and their corresponding attributes are listed in the supplementary materials. The data in Stage 1 primarily comes from the MS-COCO dataset \cite{lin2014microsoft} due to its segmentation and bounding box annotations. These annotations are essential for generating VPs and their associated question-answer pairs to assess the capabilities of MLLMs.

To define the properties of each VP shape, we begin with a detailed breakdown. A \textbf{tag} has attributes including: (1) an alphabet or digit label at its center, with a circular or square shape, and (2) font size specifying the label’s dimensions. A \textbf{bounding box} is defined by: (1) line width, representing the thickness of its outline, and (2) vertex shape, such as small squares or circles at the corners. An \textbf{arrow}, pointing from a selected direction to the target, has: (1) line width, and (2) pointer shape, which could be triangular or wedge-shaped. A \textbf{mask} is a filled region indicating the target, possibly accompanied by an outline, and consists of: (1) line width, and (2) style, which could either be a filled region with or without an outline. A \textbf{contour} captures the target's outline, characterized by: (1) line width, and (2) style, which may be precise or hand-drawn. An \textbf{oval} is an elliptical shape encircling the target, defined by the line width. A \textbf{point} is a small square or circular dot located at the target’s center, described by: (1) point size (the area it occupies), and (2) point shape (square or circle). Finally, a \textbf{scribble} is a free-form line used to indicate a target, defined by the line width of the scribble.

To evaluate the effect of VPs on visual perception, fine-grained multiple-choice questions were generated for each VP type, covering presence, location, enumeration, and referring objects, with up to four answer options depending on the question type. First, question templates were created manually with guidance from GPT-4o~\cite{OpenAI2024GPT4o}, and annotated answers from metadata were inserted into these templates. Second, distractors were produced either through manually crafted rules or by employing Qwen2-VL-72B~\cite{wang2024qwen2} with carefully designed prompts to ensure plausibility and quality. For instance, in “reference” questions, Qwen2-VL-72B generated realistic but incorrect options, whereas for “VP spatial location” questions, distractors were sampled based on randomized canvas positions. To enhance evaluation robustness, one debiased sample was included for every six entries. A debiased question consists of a visual image that excludes the VP referenced in the question, together with its corresponding QA pair, as illustrated by the \textit{debias} example in the Stage~1 evaluation in Figure~\ref{fig:dataset-overview}.

Overall, the Stage~1 data curation process establishes a comprehensive foundation for evaluating MLLMs’ VP perception capabilities in a controlled setting. By systematically defining VP shapes, attributes, and associated QA pairs, this stage isolates the model’s ability to interpret visual cues independent of downstream reasoning tasks. The resulting dataset enables fine-grained benchmarking of core VP competencies and serves as the basis for the Stage~2 evaluation, where VP understanding is integrated into complex, task-oriented scenarios.

\subsection{Stage 2 Data Curation Process} \label{sec:stage2-curation}

Stage 1 evaluates the performance of different VP types and models in natural scenes. Following this, a new question arises: Can VPs contribute to a broader range of VP-related downstream tasks? How well do existing models perform in this regard? Hence, in Stage 2 evaluation, we carefully select several widely used application scenarios, such as natural recognition, medical image analysis, human facial recognition, street view recognition, traffic sign recognition, visual relation reasoning, GUI recognition, and 3D understanding, which are further categorized into 6 tasks. To investigate whether VPs enhance MLLMs' regional perception under downstream task scenarios, we evaluate both open-source MLLMs and proprietary systems, using each model’s best-performing VP from Stage 1 results. 

For data curation, the target scenarios were first defined, followed by the specification of downstream tasks for each scenario. Relevant datasets were collected through searches on Google, Papers with Code, and Kaggle, and each dataset’s suitability and relevance were systematically evaluated. The data were then organized into a standardized metadata format that includes task descriptions, questions, answers, input contexts, regional annotations, and images. This standardized format facilitated the construction of visual question–answer pairs, and the accuracy of the information was manually verified to ensure its compatibility with multiple-choice question generation. To maintain efficiency, each task was limited to a maximum of 200 randomly selected images with relevant entries, unless the dataset contained fewer samples. A detailed description of the metadata format is provided in the supplementary materials. The final collection comprises datasets including SZ-CXR~\cite{sergii2018chest}, Gleason2019~\cite{nir2018automatic}, SD-100~\cite{li2024spatial}, Emotic~\cite{kosti2020emotic}, MapillaryVistas~\cite{neuhold2017mapillary}, SeeClick~\cite{cheng2024seeclick}, and PSG~\cite{yang2022panoptic}. These were grouped into six downstream tasks: Medical Image Analysis (MIA) using SZ-CXR and Gleason2019; 3D object recognition using SD-100; facial emotion recognition using Emotic; street view recognition using MapillaryVistas; GUI element recognition using SeeClick; and scene graph generation (SGG) using PSG.

For question and answer generation, we adapt multiple-choice visual questions (with up to four options), drawing from each sample’s metadata. We either craft rules manually or use Qwen2-VL-72B \cite{wang2024qwen2} with carefully designed prompts to ensure efficient and high-quality generation. For instance, in 3D question-answering tasks, Qwen2-VL-72B generates plausible but incorrect distractors based on the question and the correct answer, while in visual relation reasoning tasks, we randomly select misleading item classes from the metadata as alternative options.

In summary, the Stage~2 dataset extends the evaluation of VPs beyond controlled natural scenes to diverse, task-driven scenarios. By leveraging the best-performing VP configurations from Stage~1, this stage enables a systematic assessment of how visual prompting contributes to regional perception and task-specific reasoning across multiple application domains. The curated datasets and standardized question–answer pairs provide a robust foundation for benchmarking MLLMs in realistic downstream contexts, thereby bridging the gap between VP perception and practical deployment.

\begin{table*}[htbp]
  \centering
  \resizebox{0.9\textwidth}{!}{%
  \begin{tabular}{l*{8}{c}*{4}{c}c}
    \toprule
    \multirow{2}{*}{Model} &
      \multicolumn{8}{c}{\textbf{Visual Prompt Types}} &
      \multicolumn{4}{c}{\textbf{Problem Types}} &
      \multirow{2}{*}{Avg.} \\
    \cmidrule(lr){2-9}\cmidrule(lr){10-13}
      & Tag & Arrow & BBox & Contour & Mask & Oval & Point & Scribble &
      Enumeration & Existence & Rough-Loc. & VP-Ref. \\
    \midrule
\multicolumn{14}{l}{\textit{Human Baseline}} \\
Human Reviewers & 89.00 & 92.62 & 97.29 & 87.68 & 85.28 & 94.87 & 90.68 & 82.80 & 90.73 & 94.36 & 97.68 & 84.26 & 90.03 \\
\midrule
\multicolumn{14}{l}{\textit{Proprietary Models}} \\
GPT-4o & 69.95 & 70.27 & 74.18 & 77.18 & 65.32 & 79.77 & 49.28 & 64.45 & 60.44 & 87.03 & 57.83 & 67.74 & 68.80 \\
Doubao-Seed-1.6 & 86.21 & 79.60 & 93.60 & 72.43 & 71.19 & 50.17 & 47.49 & 62.30 & 62.47 & 64.04 & \underline{95.34} & \textbf{80.53} & 70.37 \\
Qwen-VL-Max & 92.27 & 82.84 & 93.10 & 88.75 & 67.51 & 92.84 & 69.31 & 74.45 & 81.80 & 88.11 & 92.13 & 76.82 & 82.63 \\
\midrule
\multicolumn{14}{l}{\textit{Pre-trained Models}} \\
LLaVA-1.5-7B & 67.41 & 66.75 & 69.29 & 64.25 & 67.35 & 67.72 & 59.45 & 49.36 & 62.92 & 96.69 & 45.54 & 53.28 & 63.95 \\
CogVLM2-LLama3-Chat-19B & 74.54 & 69.96 & 79.56 & 75.01 & 48.22 & 81.54 & 40.75 & 43.47 & 71.87 & 57.97 & 73.26 & 64.13 & 64.88 \\
InternVL3-1B & 74.99 & 79.91 & 78.59 & 75.51 & 56.00 & 84.06 & 58.68 & 63.93 & 79.56 & 67.49 & 78.83 & 63.45 & 71.46 \\
Qwen2.5-VL-3B-Instruct & 87.16 & 79.91 & 85.41 & 80.18 & 52.18 & 84.58 & 55.76 & 64.10 & 73.07 & 76.89 & 89.20 & 65.29 & 73.41 \\
LLaVA-1.5-13B & 77.98 & 73.48 & 77.80 & 78.47 & 75.90 & 76.78 & 73.17 & 68.82 & 83.17 & \textbf{98.77} & 60.17 & 60.58 & 75.47 \\
Llama-3.2-90B-Vision-Instruct & 69.17 & 79.71 & 91.49 & 88.71 & 74.28 & 91.69 & 59.28 & 71.79 & 77.89 & 81.51 & 79.23 & 71.76 & 78.45 \\
DeepSeek-VL2 & 87.36 & 78.07 & 80.62 & 82.24 & 72.44 & 75.76 & 76.80 & 66.20 & 73.92 & 91.41 & 80.86 & 71.37 & 77.94 \\
LLaVA-OneVision-Qwen2-7B-OV & 89.36 & 81.67 & 88.09 & 88.89 & 66.97 & 92.89 & 60.76 & 69.99 & 79.80 & 84.99 & 91.47 & 69.43 & 79.58 \\
InternVL3-2B & 89.17 & 84.73 & 88.92 & 87.46 & 57.15 & 87.79 & 79.08 & 69.87 & 81.41 & 88.12 & 89.70 & 70.23 & 82.27 \\
GLM-4V-9B & 85.43 & 79.62 & 87.17 & 86.65 & 74.00 & 85.84 & 75.46 & 74.71 & \textbf{91.37} & 91.44 & 70.77 & 74.20 & 82.36 \\
Qwen2.5-VL-7B-Instruct & 90.39 & 82.94 & 91.19 & 88.25 & 66.51 & 91.51 & 67.41 & 73.08 & 83.47 & 86.81 & 92.10 & 70.17 & 81.29 \\
Qwen2.5-VL-32B-Instruct & 91.11 & 82.97 & 92.04 & 87.41 & 69.29 & 93.64 & 75.29 & 74.13 & 83.94 & 90.13 & 91.31 & 73.49 & 83.22 \\
InternVL3-14B & 91.14 & 85.17 & 93.03 & 90.42 & 70.47 & \underline{94.67} & 73.83 & 69.64 & 83.18 & 89.73 & 93.51 & 74.75 & 83.54 \\
Qwen2.5-VL-72B-Instruct & 92.26 & 82.57 & 92.88 & 88.62 & 68.68 & 92.83 & 69.58 & 74.77 & 81.84 & 88.26 & 92.01 & 75.79 & 82.80 \\
Ovis2-34B & 91.86 & 84.39 & 93.03 & 90.63 & 69.49 & 94.38 & 72.09 & 72.49 & 84.72 & 88.19 & 92.49 & 75.59 & 83.24 \\
InternVL3-8B & 91.25 & 84.94 & 92.86 & \textbf{92.83} & 71.60 & 93.52 & 70.52 & 75.35 & 86.68 & 89.13 & 94.19 & 72.50 & 84.11 \\
LLaVA-v1.6-34B-HF & 90.89 & 86.00 & 86.46 & 86.59 & 77.61 & 91.11 & \textbf{84.28} & 76.86 & 82.62 & 94.97 & 93.86 & 72.24 & 84.97 \\
InternVL3-9B & 92.19 & \underline{86.81} & 93.76 & 90.68 & 73.01 & 94.21 & 70.61 & 78.73 & 86.51 & 88.12 & 93.41 & 77.37 & 85.00 \\
NVLM-D-72B & 91.28 & \textbf{88.23} & 91.60 & 90.22 & 76.02 & 93.35 & 77.00 & 75.41 & 83.82 & 93.99 & 92.06 & 76.50 & 85.39 \\
Molmo-72B-0924 & 90.71 & 85.31 & 92.92 & 89.61 & 79.05 & 93.18 & 76.57 & 77.56 & 87.18 & \underline{97.24} & 92.46 & 70.08 & 85.61 \\
InternVL3-38B & \underline{92.73} & 86.49 & \textbf{94.33} & \underline{92.28} & 77.15 & 93.86 & 78.88 & \underline{79.37} & 87.84 & 91.61 & \textbf{95.37} & 77.93 & \underline{86.89} \\
InternVL3-78B & \textbf{93.87} & 85.77 & \underline{94.25} & 91.56 & \underline{80.01} & \textbf{95.81} & \underline{81.59} & \textbf{80.89} & \underline{88.59} & 92.93 & 95.24 & \underline{79.65} & \textbf{87.97} \\
\midrule
\multicolumn{14}{l}{\textit{Fine-tuned with Visual-Prompt Data}} \\
ViPLLaVA-13B & 75.00 & 53.02 & 84.37 & 86.27 & \textbf{83.79} & 56.31 & 46.66 & 59.50 & 80.36 & 61.43 & 87.70 & 57.28 & 68.12 \\
ViPLLaVA-7B & 76.44 & 69.78 & 81.71 & 83.78 & 75.47 & 65.77 & 65.38 & 62.30 & 77.51 & 84.84 & 83.01 & 53.58 & 72.58 \\
\midrule
\textbf{Average} & 85.26 & 79.66 & 87.49 & 85.00 & 69.88 & 85.02 & 67.22 & 69.39 & 79.92 & 85.62 & 84.03 & 70.21 & 78.61 \\
    \bottomrule
  \end{tabular}}%
  \caption{Performance on VP‐shape and question‐type subtasks, sorted by overall accuracy. Best of each task excluding human baseline is \textbf{bold} and second-best is \underline{underlined}.}
  \label{tab:model_comparison}
\end{table*}

\section{Experiment}
\label{sec:exp}

\subsection{Experiment Setup}
\label{sec:exp-setup}
\textbf{Compared Models}. In this study, we selected 3 proprietary models along with 24 open-source MLLMs across various categories. These include popular visual models (e.g., Qwen2.5-VL, InternVL-3, and Llama-3.2-Vision), specialized models (e.g., MiniCPM-V 2.6, CogVLM2, and GLM-4V), and recently introduced vision-language architectures (e.g., NVLM, DeepSeek-VL2, Ovis2, and Molmo). To further illustrate the impact of model parameter scale and architecture on visual understanding, we incorporated models of different scales from the InternVL3 and Qwen2.5-VL families.

\noindent
\textbf{Evaluation Metrics}. Our proposed VP-Bench is comprised of two stages, both stages are in multi-choice question format, e.g., ``Where is VP located? Options: (A) In the bottom, (B) In the top". Generally, we follow the VLMEvalKit~\cite{duan2024vlmevalkit} procedure to evaluate models' performance. Accuracy is the primary metric.

\begin{table*}[ht]
  \centering
  \resizebox{\textwidth}{!}{%
  \begin{tabular}{l l *{14}{r}}
    \toprule
    \multirow{2}{*}{\textbf{Model}} &
      \multirow{2}{*}{\textbf{Best Attribution Combination}} &
      \multicolumn{2}{c}{\textbf{MIA}} &
      \multicolumn{2}{c}{\textbf{SD-100}} &
      \multicolumn{2}{c}{\textbf{Emotic}} &
      \multicolumn{2}{c}{\textbf{Mapillary Vistas}} &
      \multicolumn{2}{c}{\textbf{SeeClick}} &
      \multicolumn{2}{c}{\textbf{SGG}} &
      \multicolumn{2}{c}{\textbf{Average}} \\
    \cmidrule(lr){3-4}\cmidrule(lr){5-6}\cmidrule(lr){7-8}%
    \cmidrule(lr){9-10}\cmidrule(lr){11-12}\cmidrule(lr){13-14}\cmidrule(lr){15-16}
      & & R-BVP & BVP\,($\Delta$) & R-BVP & BVP\,($\Delta$) & R-BVP & BVP\,($\Delta$) &
        R-BVP & BVP\,($\Delta$) & R-BVP & BVP\,($\Delta$) & R-BVP & BVP\,($\Delta$) &
        R-BVP & BVP\,($\Delta$) \\
    \midrule
    \multicolumn{16}{l}{\textit{Proprietary Models}} \\
Doubao-Seed-1.6 & bbox(contrast, round, thick)
        & \underline{57.56} & 59.60 (+2.04) & \textbf{87.00} & \textbf{89.33} (+2.33) & \underline{72.95} & 70.48 (-2.47)
        & \underline{78.54} & 76.08 (-2.46) & \underline{98.01} & 97.22 (-0.79) & 95.91 & \textbf{96.18} (+0.27) & 81.66 & 81.48 (-0.18)\\
GPT-4o & Tag(digit, blue, round, small)
        & 54.25 & \underline{60.60} (+6.35) & 74.00 & 72.24 (-1.76) & 65.61 & 66.04 (+0.43)
        & 68.78 & 57.89 (-10.89) & 97.44 & 97.00 (-0.44) & 81.79 & 81.60 (-0.19) & 73.64 & 72.56 (-1.08)\\
Qwen-VL-Max & bbox(contrast, none, medium)
        & 42.83 & 44.32 (+1.49) & 80.67 & 81.27 (+0.60) & 74.02 & 73.03 (-0.99)
        & 64.39 & 62.68 (-1.71) & 96.30 & 96.89 (+0.59) & \underline{95.37} & 95.44 (+0.07) & 75.60 & 75.61 (+0.01)\\
    \midrule
    \multicolumn{16}{l}{\textit{Pre-trained Models}} \\
CogVLM2-LLama3-Chat-19B & oval(contrast, thin)
        &  4.20 &  5.40 (+1.20) & 27.67 & 31.33 (+3.66) & \underline{74.40} & 70.71 (-3.69)
        & 29.27 & 33.49 (+4.22) & 70.17 & 75.31 (+5.14) & 84.36 & 83.30 (-1.06) & 48.35 & 49.92 (+1.57)\\
DeepSeek-VL2 & tag(alphabet, green, round, large)
        & 19.22 & 20.90 (+1.68) & 24.75 & 23.00 (-1.75) & 66.18 & 63.76 (-2.42)
        & 36.59 & 24.88 (-11.71) & 86.93 & 67.50 (-19.43) & 85.20 & 83.30 (-1.90) & 53.15 & 47.22 (-5.93)\\
GLM-4V-9B & contour(contrast, contour, thick)
        & 19.12 & 18.19 (-0.93) & 61.00 & 59.33 (-1.67) & 66.67 & 64.94 (-1.73)
        & 43.90 & 53.59 (+9.69) & 96.59 & 96.25 (-0.34) & 88.62 & 86.32 (-2.30) & 62.65 & 63.10 (+0.45)\\
InternVL3-1B & oval(contrast, thin)
        & 19.72 & 17.20 (-2.52) & 47.00 & 46.67 (-0.33) & 57.25 & 47.14 (-10.11)
        & 40.98 & 43.54 (+2.56) & 91.48 & 92.90 (+1.42) & 88.89 & 88.45 (-0.44) & 57.55 & 55.98 (-1.57)\\
InternVL3-2B & bbox(contrast, none, thick)
        & 38.64 & 40.40 (+1.76) & 46.67 & 51.33 (+4.66) & 66.18 & \textbf{66.67} (+0.49)
        & 47.32 & 54.07 (+6.75) & 94.60 & 98.46 (+3.86) & 89.96 & 90.23 (+0.27) & 63.89 & 66.86 (+2.97)\\
InternVL3-8B & bbox(blue, round, medium)
        & \underline{45.55} & 46.60 (+1.05) & 65.67 & 75.00 (+9.33) & \textbf{74.64} & 71.76 (-2.88)
        & 49.76 & 52.63 (+2.87) & \underline{97.73} & 97.50 (-0.23) & 92.62 & 93.87 (+1.25) & 71.00 & 72.89 (+1.89)\\
InternVL3-9B & bbox(contrast, square, thin)
        & 47.75 & 48.90 (+1.15) & 69.33 & 76.33 (+7.00) & 73.43 & \textbf{74.29} (+0.86)
        & 56.59 & 61.72 (+5.13) & 96.31 & 98.15 (+1.84) & 90.93 & 91.83 (+0.90) & 72.39 & 75.20 (+2.81)\\
InternVL3-14B & bbox(contrast, square, thin)
        & 30.83 & 29.90 (-0.93) & 71.33 & 77.33 (+6.00) & \underline{74.40} & 72.86 (-1.54)
        & \textbf{58.54} & 61.72 (+3.18) & 97.16 & 98.77 (+1.61) & 92.62 & 93.07 (+0.45) & 70.81 & 72.27 (+1.46)\\
InternVL3-38B & bbox(red, square, thick)
        & 46.65 & 48.40 (+1.75) & \underline{85.00} & 88.67 (+3.67) & 71.98 & 72.94 (+0.96)
        & 59.02 & \underline{59.81} (+0.79) & \underline{98.58} & 99.00 (+0.42) & \underline{93.24} & \underline{94.23} (+0.99) & 75.74 & 77.17 (+1.43)\\
InternVL3-78B & bbox(red, none, medium)
        & 47.55 & \textbf{53.40} (+5.85) & \textbf{86.00} & 87.00 (+1.00) & 72.95 & \underline{73.65} (+0.70)
        & \underline{65.37} & \underline{67.46} (+2.09) & 98.01 & 97.75 (-0.26) & \textbf{94.67} & 95.03 (+0.36) & \textbf{77.42} & \textbf{79.05} (+1.63)\\
Llama-3.2-90B-Vision-Instr. & bbox(green, square, medium)
        & 54.75 & 59.50 (+4.75) & 60.00 & 58.00 (-2.00) & 61.59 & 65.65 (+4.06)
        & 46.83 & 51.20 (+4.37) & 97.44 & 98.50 (+1.06) & 86.13 & 88.99 (+2.86) & 67.79 & 70.31 (+2.52)\\
LLaVA-v1.5-7B & tag(digit, red, square, medium)
        & 29.53 & 35.30 (+5.77) & 29.67 & 31.00 (+1.33) & 41.06 & 33.65 (-7.41)
        & 11.71 & 13.88 (+2.17) & 82.39 & 83.50 (+1.11) & 83.47 & 73.87 (-9.60) & 46.31 & 45.20 (-1.11)\\
LLaVA-v1.5-13B & tag(alphabet, red, square, large)
        & 37.94 & 30.60 (-7.34) & 27.67 & 24.00 (-3.67) & 49.52 & 39.76 (-9.76)
        & 19.51 & 17.22 (-2.29) & 84.66 & 80.25 (-4.41) & 87.02 & 83.30 (-3.72) & 51.05 & 45.85 (-5.20)\\
LLaVA-OneVision-Qwen2-7B-OV-HF & oval(blue, thick)
        & 18.02 & 16.72 (-1.30) & 60.33 & 63.67 (+3.34) & 65.13 & 64.39 (-0.74)
        & 41.46 & 47.37 (+5.91) & 97.16 & 98.50 (+1.34) & 91.56 & 92.01 (+0.45) & 62.28 & 63.78 (+1.5)\\
LLaVA-v1.6-34B-HF & oval(contrast, thin)
        & 39.84 & 37.60 (-2.24) & 48.00 & 52.33 (+4.33) & 68.12 & 64.05 (-4.07)
        & 42.93 & 48.33 (+5.40) & 94.03 & 96.30 (+2.27) & 92.09 & 91.30 (-0.79) & 64.17 & 64.98 (+0.81)\\
MiniCPM-V-2\_6 & tag(digit, red, square, large)
        & 14.61 & 15.30 (+0.69) & 55.00 & 38.00 (-17.00) & 72.22 & 67.06 (-5.16)
        & 40.49 & 26.32 (-14.17) & 96.02 & 92.75 (-3.27) & 89.96 & 89.08 (-0.88) & 61.38 & 54.75 (-6.63)\\
Molmo-72B-0924 & bbox(contrast, square, thin)
        & \textbf{61.46} & \textbf{62.80} (+1.34) & 78.00 & 78.00 (+0.00) & 67.63 & 67.38 (-0.25)
        & 63.41 & 60.77 (-2.64) & 95.17 & 96.30 (+1.13) & 90.93 & 91.83 (+0.90) & \underline{76.10} & \underline{76.18} (+0.08)\\
NVLM-D-72B & bbox(red, square, thick)
        & 53.45 & 58.10 (+4.65) & 71.33 & 77.00 (+5.67) & 69.08 & 72.71 (+3.63)
        & 51.22 & 49.76 (-1.46) & 94.60 & 97.00 (+2.40) & 93.69 & 93.43 (-0.26) & 72.23 & 74.67 (2.44)\\
Ovis2-34B & bbox(contrast, square, thin)
        & 33.93 & 37.30 (+3.37) & 78.33 & 83.33 (+5.00) & \underline{74.40} & 73.10 (-1.30)
        & 53.66 & 50.72 (-2.94) & \textbf{99.15} & \textbf{99.69} (+0.54) & \textbf{93.69} & \textbf{94.58} (+0.89) & 72.19 & 73.12 (+0.93)\\
Qwen2.5-VL-3B-Instr. & bbox(red, none, medium)
        & 22.92 & 23.80 (+0.88) & 68.67 & 66.00 (-2.67) & 68.84 & 68.24 (-0.60)
        & 39.02 & 44.02 (+5.00) & 95.74 & 98.00 (+2.26) & 94.49 & 94.49 (+0.00) & 64.95 & 65.76 (+0.81)\\
Qwen2.5-VL-7B-Instr. & bbox(contrast, square, medium)
        & 22.12 & 22.50 (+0.38) & 76.00 & 77.33 (+1.33) & 69.32 & 70.71 (+1.39)
        & 43.90 & 52.63 (+8.73) & 96.59 & \underline{99.07} (+2.48) & 92.62 & 92.27 (-0.35) & 66.76 & 69.08 (+2.32)\\
Qwen2.5-VL-32B-Instr. & bbox(contrast, none, medium)
        & 32.00 & 39.60 (+7.60) & 75.33 & 80.00 (+4.67) & 65.71 & 66.43 (+0.72)
        & 53.85 & 47.37 (-6.48) & \underline{97.73} & 97.84 (+0.11) & 93.24 & 94.32 (+1.08) & 69.64 & 70.93 (+1.29)\\
Qwen2.5-VL-72B-Instr. & bbox(contrast, square, thin)
        & 42.44 & 43.80 (+1.36) & 81.67 & 83.33 (+1.66) & 73.91 & 71.43 (-2.48)
        & 64.88 & 61.72 (-3.16) & 96.59 & 97.53 (+0.94) & \underline{95.20} & \underline{95.47} (+0.27) & \underline{75.78} & \underline{75.55} (-0.23)\\
    \midrule
    \multicolumn{16}{l}{\textit{Fine-tuned with Visual-Prompt Data}} \\
ViP-LLaVA-7B & contour(blue, contour, medium)
        & 32.73 & 36.80 (+4.07) & 37.33 & 36.33 (-1.00) & 48.07 & 46.59 (-1.48)
        & 25.85 & 11.00 (-14.85) & 82.39 & 80.25 (-2.14) & 87.56 & 83.84 (-3.72) & 52.32 & 49.13 (-3.19)\\
ViP-LLaVA-13B & mask(red, fill, medium)
        & 31.53 & 33.60 (+2.07) & 29.33 & 28.00 (-1.33) & 58.21 & 59.53 (+1.32)
        & 27.32 & 33.97 (+6.65) & 84.66 & 82.50 (-2.16) & 87.29 & 88.72 (+1.43) & 53.06 & 54.39 (+1.33)\\
    \midrule
    \textbf{Average} & & 35.75 & 37.39 (+1.64) & 60.81 & 61.96 (+1.15) & 66.55 & 64.96 (-1.59) & 47.32 & 47.35 (0.02) & 93.34 & 93.23 (-0.11) & 90.46 & 90.01 (-0.45) & & \\
    \bottomrule
  \end{tabular}}%
  \caption{Performance comparison on VP‐related tasks. “Best VP Attr.\ Comb.” indicates the Stage 1 attribute combination. Best in each task is \textbf{bold} and second-best is \underline{underlined}.}
  \label{tab:visual_recognition_comparison}
\end{table*}

\subsection{Evaluation Main Results}
This section evaluates MLLMs on VP-Bench alongside Human baselines. We report the overall score for all perceptual tasks in Table~\ref{tab:model_comparison} as well as the best performance on each downstream task in Table~\ref{tab:visual_recognition_comparison}. Various instruction arrangements for all tasks are investigated. We summarize the key findings as follows.

In Stage 1 evaluation, we present the average accuracy of all models across eight VP shapes and four question types in Table~\ref{tab:model_comparison}, along with the human baseline. ``Avg." denotes the average accuracy across all QA samples. In terms of overall accuracy, InternVL3-78, InternVL3-38B, and Molmo-72B rank among the top three, with accuracy around 87\%. Other statistics are as follows.

Across all \textbf{question types}, most models accurately detect, count, and localize VP, achieving over 90\% accuracy on existence queries, around 85\% on enumeration queries, and over 92\% on rough-location queries. Their performance on referring-expression resolution remains substantially lower: mean accuracy hovers around 70\%, and the community favored Qwen-2.5-VL-72B reaches only 75.79\%. When benchmarked against human annotators, MLLMs still exhibit an around 10\% deficit.

An appropriate \textbf{VP shape} can significantly enhance a model's ability to detect both the prompt and the highlighted region. For example, the bounding box shape is generally a better comparison to the point shape in Table~\ref{tab:model_comparison}, as models can achieve an average of 87.49\% with the bounding box shape, but only 67.22\% accuracy with the point shape. As well as reflecting most of the models' best recognized \textbf{VP attributes} combination, there are variations of the bounding box shape in Table~\ref{tab:visual_recognition_comparison}. The results indicate that a contrast color bounding box with medium thickness is optimal for more than half of the models, while a contrast color oval with thin edges is also the most effective for many models, where the contrast color is selected as one of the most distinctive red, green, or blue hues relative to the background. Overall contrast color emerges as the preferred choice for most models. This suggests that color plays a crucial role in distinguishing the region of interest from the background. Regarding scale, thin to medium scale VP yields better results, implying that MLLMs exhibit greater perceptual sensitivity to thin prompts while preserving more contextual information.

\subsection{Results Analysis}
To further interrogate our preliminary findings, we carried out a series of supplementary experiments whose results yielded several salient statistical patterns. These additional tests not only corroborate the baseline trends but also uncover nuanced insights into the models’ behavior under varying prompt conditions. The remainder of this section highlights the principal observations before delving into a detailed, case-by-case analysis.

\textbf{Shape regular VPs are more readily perceived by MLLMs than shape irregular ones}. As shown in Table~\ref{tab:model_comparison}, models recognize regular VPs like Tag, Arrow, Bounding Box, Oval, at around 80\% accuracy on average, whereas for irregular VPs like Mask, Point, Scribble, they only reach less than 70\%. Moreover, the gap relative to the human baseline is larger on those irregular shapes. A per‐model breakdown shows that InternVL3-78B, while still 9.79\% behind humans on Point, differs by only about 2\% on Mask and Scribble, and even exceeds human performance on irregular contours (91.56\% compared to 87.68\%). Notably, DeepSeek-VL2 and InternVL3-14B suffer their worst scores on hand-drawn scribbles. This pattern suggests that training data are biased toward regular geometric forms, with fewer examples of irregular shapes, which in turn impairs MLLMs’ ability to detect those VP forms.


\begin{table*}[t]
\centering
\small
\begin{tabular}{lccccccccc}
\toprule
Model & Tag & Arrow & BBox & Contour & Mask & Oval & Point & Scribble & Avg. \\
\midrule
\textit{InternVL3-78B} & & & & & & & & & \\
w/o VP description
 & 91.69 & 85.45 & 94.06 & 90.19 & 50.77 & 92.49 & 69.21 & 76.52 & 81.30 \\
w. VP description
 & 93.87
 & 85.77
 & 94.25
 & 91.56
 & 80.01
 & 95.81
 & 81.59
 & 80.89
 & 87.97 \\
\midrule
\textit{Qwen2.5-VL-72B-Instruct} & & & & & & & & & \\
w/o VP description
 & 85.93 & 81.96 & 92.88 & 87.81 & 41.38 & 82.80 & 72.63 & 69.76 & 76.89 \\
w. VP description
 & 92.26
 & 82.57
 & 92.88
 & 88.62
 & 68.68
 & 92.83
 & 69.58
 & 74.77
 & 82.77 \\
\bottomrule
\end{tabular}
\caption{Comparison of model performance with and without VP descriptions on Stage 1 evaluation. \textbf{Tag} is the mean of Alphabet and Digit. \textbf{Avg.} is the mean over \{Tag, Arrow, BBox, Contour, Mask, Oval, Point, Scribble\}.}
\label{tab:model_context_comparison}
\end{table*}

\textbf{An explicit description of VP shapes is critical for enabling MLLMs to interpret them in context}. To examine whether models truly understand VPs, we compared two settings: with and without a VP‐shape description inserted into the instruction. As shown in Table~\ref{tab:model_context_comparison}, experiments were conducted on InternVL3-78B and Qwen2.5-VL-72B, incorporating VP descriptions consistently enhanced the models’ comprehension of VPs, though the effect varied across VP types. For instance, when VP descriptions were included in the \textbf{Arrow} and \textbf{Contour} scenarios, InternVL3-78B achieved only marginal improvements of 0.32\% and 1.37\%, respectively, while Qwen2.5-VL-72B improved by just 0.61\% and 0.81\%. In contrast, substantial gains were observed in the \textbf{Mask} and \textbf{Point} scenarios: InternVL3-78B improved by 29.24\% and 12.38\%, respectively, whereas Qwen2.5-VL-72B improved by 27.3\% in the \textbf{Mask} scenario but declined by 3.05\% in the \textbf{Point} scenario. These differences are likely influenced by the distribution of training data, where the frequency of each VP type affects the extent to which descriptive prompts enhance model performance. Collectively, these results suggest that providing an explanatory language description of the VP can help models more accurately identify and interpret the corresponding visual cue in the image.

\textbf{VPs that a model perceives most accurately are generally the best choices for downstream tasks but not always}. To investigate this, we devised two selection schemes:
\begin{enumerate}
  \item \textbf{Random Best VP (R-BVP):} is randomly selected from the set of VP attribute combinations that achieved the highest perception performance across all models in the Stage~1 evaluation.
  \item \textbf{Best VP (BVP):} always uses the single VP attribute combination the current model perceives most accurately.
\end{enumerate}

As shown in Table~\ref{tab:visual_recognition_comparison}, for Qwen-2.5-VL-72B, the performance gap between BVP and R-BVP stays within $\pm$ 5\%, and in over half of the tasks BVP outperforms R-BVP. Likewise, InternVL3-78B shows slightly better results with BVP in five tasks, for example, a +5.85\% gain on MIA and +1.00\% on SD-100, and only trails by 0.26\% on SeeClick. By contrast, DeepSeek-VL2 exhibits much larger disparities: in most downstream evaluations, BVP underperforms R-BVP (–11.71\% on MapillaryVistas and –19.43\% on SeeClick), yet it exceeds R-BVP by +1.68\% on MIA. A similar pattern appears with GPT-4o: BVP is +6.35\% better on MIA but –10.89\% worse on MapillaryVistas. These findings suggest that, although a model’s perception accuracy is generally the main criterion for choosing a VP, a robust VP-selection strategy is also critical for maximizing downstream performance.  

\textbf{Simply training models with VP data offers no clear benefit}. 
ViP-LLaVA extends the LLaVA architecture by incorporating VP-related data during the instruction tuning stage to enhance VP perception, while LLaVA-1.5 serves as the baseline model in our experiments for comparison against ViP-LLaVA trained with VP-enriched datasets. In Stage~1, LLaVA-1.5-7B achieved an average accuracy of 64.33\%, whereas ViP-LLaVA-7B reached 73.01\%. However, in Stage~2 downstream tasks, ViP-LLaVA-7B underperformed LLaVA-1.5-7B by 2.88\% on MapillaryVistas and by 3.25\% on SeeClick. Furthermore, this straightforward training approach led to degradation at larger scales: ViP-LLaVA-13B’s Stage~1 accuracy dropped by 4.13\% relative to ViP-LLaVA-7B and by 6.72\% relative to LLaVA-1.5-13B. These findings indicate that downstream performance depends more on a model’s robust foundational capabilities and that improving VP perception without compromising these core abilities requires more balanced data composition and refined training strategies.

\section{Conclusion}
\label{sec:conclusion}

We introduce VP-Bench, a two-stage evaluation framework for assessing the capabilities of MLLMs in perceiving VP and solving grounded referring queries. In Stage 1, we construct a dataset of over 30,000 VP images, covering 8 distinct VP shapes and 355 attribute combinations, to evaluate a model’s general understanding of VPs. Our results show that while MLLMs perform well in VP recognition and object counting, they struggle with spatial localization and fine-grained understanding. Additionally, existing models exhibit a preference for bounding boxes and tags and show heightened sensitivity to red VPs. In Stage 2, we introduce 6 VP-related downstream tasks to evaluate how well models integrate VP perception for practical problem-solving. Experimental results suggest that VPs offer certain advantages over text-based spatial prompts for these tasks. However, model performance remains largely dependent on domain knowledge. Overall, this work aims to highlight the need to refine VP attribute representations and enhance spatial reasoning, ultimately improving model interpretability and real-world applicability.

\section*{Acknowledgments}
This work is supported by the National Natural Science Foundation of China (Grant No. 62502174).

\bibliography{aaai2026}

\clearpage

\appendix
\section{Appendix}
\label{app:vpbench}

\begin{table}[ht]
    \centering
    \scriptsize
    \begin{tabular}{l}
        \hline
        \begin{minipage}[t]{\columnwidth}
\verb|[|\\[2pt]
\verb|  {|\\[2pt]
\verb|    "meta_source": "Where is this data|\\[2pt]
\verb|                   retrieved?",|\\[2pt]
\verb|    "data_file": "specify your respective separate|\\[2pt]
\verb|                 data file name",|\\[2pt]
\verb|    "region": [|\\[2pt]
\verb|      {|\\[2pt]
\verb|        "meta_bbox": ["x1", "y1", "x2", "y2"],|\\[2pt]
\verb|        "meta_polygon": "segmentation mask in|\\[2pt]
\verb|                        polygon format",|\\[2pt]
\verb|        "meta_annotation": "original data annotation",|\\[2pt]
\verb|        "question": "What does <VP> denote a region?",|\\[2pt]
\verb|        "answer": "Propose a Q-A answer."|\\[2pt]
\verb|      }|\\[2pt]
\verb|    ]|\\[2pt]
\verb|  }|\\[2pt]
\verb|]|\\[2pt]
        \end{minipage} \\
        \hline
    \end{tabular}
    \caption{The example of the metadata.}
    \label{tab:meta_example}
\end{table}



\begin{table}[h]
\centering
\small
\begin{tabular}{p{2cm}p{5cm}}
\toprule
\textbf{VP Shape} & \textbf{Context} \\
\midrule
Tag & A tag is a small label located at the center of a target, displaying a number or letter. It may be in red, blue, green, or black, and can be circular or square in shape. \\
Bounding Box & A bounding box is a rectangular frame that marks a target, which may be in red, blue, green, or black. \\
Arrow & An arrow is a symbol that points to a target, which may be in red, blue, green, or black. \\
Mask & A mask is a filled area used to indicate a target region, which may be in red, blue, green, or black. \\
Contour & A contour is the outline of a target, which may be in red, blue, green, or black. It can be drawn precisely along the outline or may resemble a loosely hand-drawn line. \\
Oval & An oval is an elliptical shape that encircles a target, which may be in red, blue, green, or black. \\
Point & A point is a square or circular dot that represents a target, located at the center of the marked target, which may be in red, blue, green, or black. \\
Scribble & A scribble is a random hand-drawn line that indicates a target, which may be in red, blue, green, or black. \\
\bottomrule
\end{tabular}
\caption{The Description of Visual Prompt Shapes.}
\label{tab:visual_prompt_shapes}
\end{table}

\section{Benchmark Curation}
\begin{figure*}[ht]
    \centering
    \includegraphics[width=0.8\linewidth]{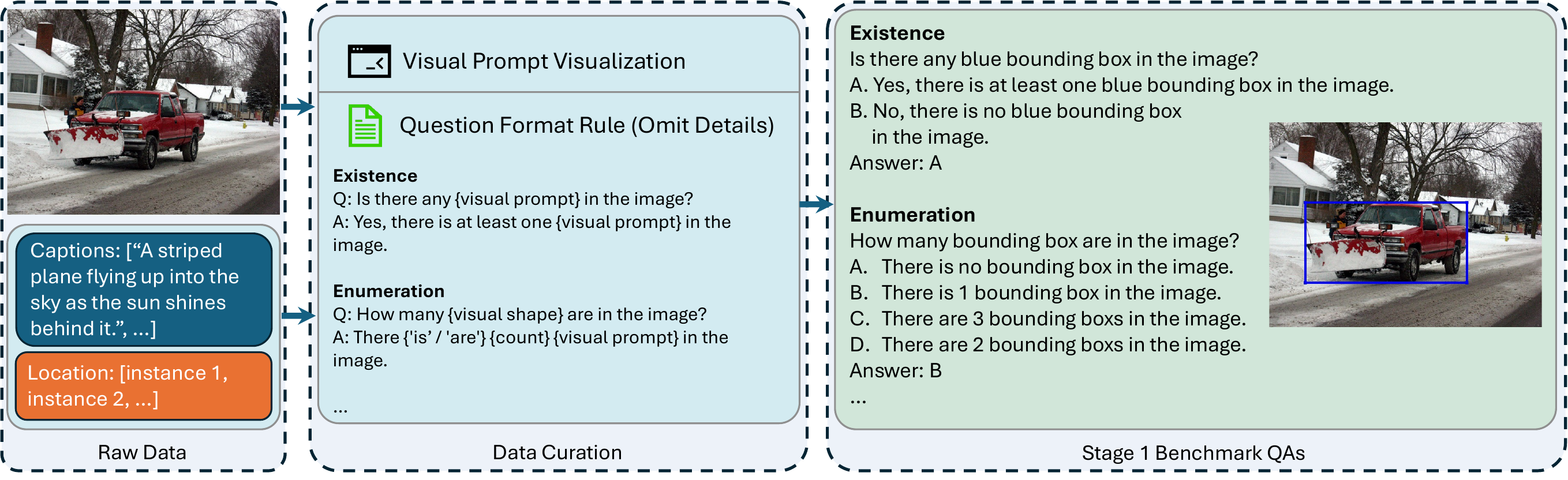}
    \caption{Benchmark Stage 1 Q\&As generation pipeline.}
    \label{fig:stage-1-gen}
\end{figure*}

\begin{figure*}[ht]
    \centering
    \includegraphics[width=0.8\linewidth]{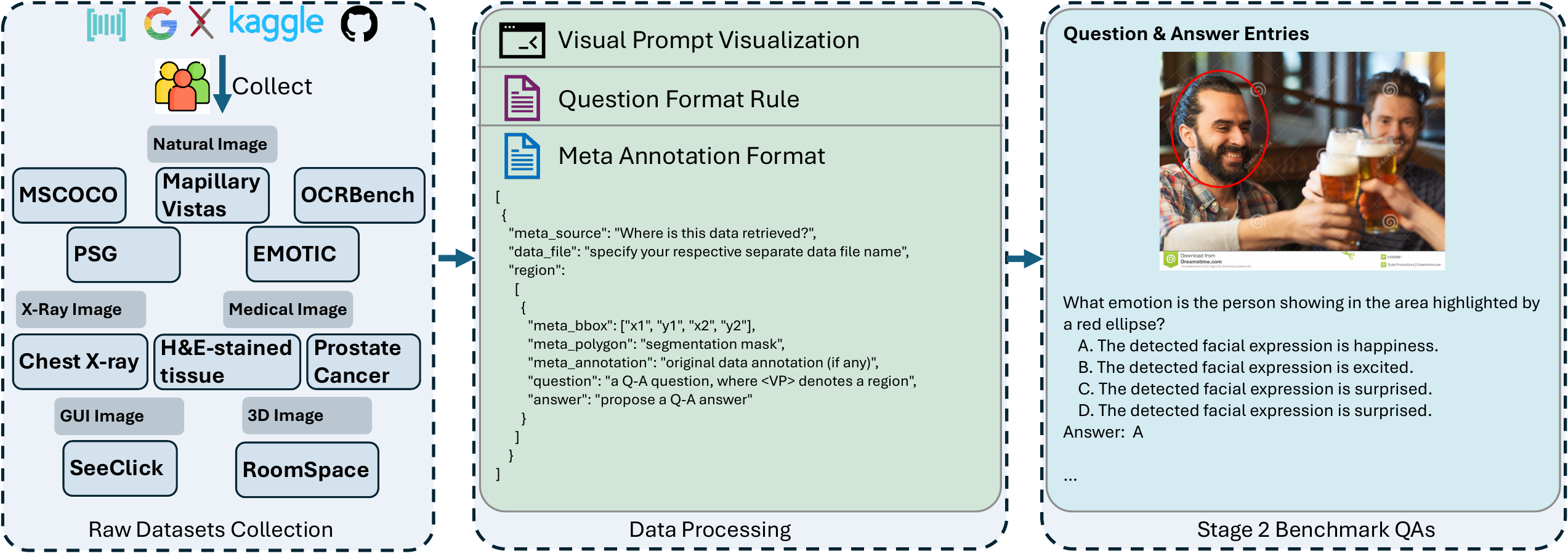}
    \caption{Benchmark Stage 2 Q\&As generation pipeline.}
    \label{fig:stage-2-gen}
\end{figure*}

We propose a benchmark consisting of two evaluation stages. To accurately achieve the goals of each stage, we collected data separately for stages 1 and 2 and defined a unified data‐generation and review pipeline as in Figures~\ref{fig:stage-1-gen} and~\ref{fig:stage-2-gen}. In both stages, these pipelines follow five phases: raw data collection, data‐quality assessment, question definition, QA generation, and visual‐prompt (VP) rendering. Below, we describe each phase in detail.

\subsection{Stage 1 Data Curation}
\noindent
\textbf{Question Definition}. The objective of Stage~1 is to evaluate how the model perceives VPs. To this end, we define four core aspects of VP perception: \textit{existence}, \textit{enumeration}, \textit{location}, and \textit{reference}. Specifically:
\begin{enumerate}
    \item \textbf{Existence}. “Does a VP of this particular color and shape appear in the image or not?”
    \item \textbf{Enumeration}. “How many VPs of this specific shape are present in the image?”
    \item \textbf{Location}. “What is the position (e.g., top-left, bottom-right) of a given VP in the image?”
    \item \textbf{Reference}. “Which object is being highlighted by the VP in the image?”
\end{enumerate}

\noindent
\textbf{Data Collection}. The dataset must support all four question types, so we chose an object-detection or semantic segmentation dataset. To minimize domain shift, letting the model focus on VP perception, we selected MS-COCO~\cite{lin2014microsoft}, which is widely used as pretraining data for multimodal LLMs.


\noindent

\begin{table}[htbp]
  \centering
  \resizebox{\linewidth}{!}{%
    \begin{tabular}{@{}ll@{}}
      \toprule
      \textbf{Question Types} & \textbf{Question Template} \\ 
      \midrule
      Existence   & Is there any \texttt{<VP>} in the image? \\ 
      Enumeration & How many \texttt{<color>} \texttt{<VP>} are in the image? \\ 
      Location    & Where is the \texttt{<color>} \texttt{<VP>} located in the image? \\ 
      Reference   & What is the \texttt{<color>} \texttt{<VP>} referring to in the image? \\ 
      \bottomrule
    \end{tabular}%
  }
  \caption{Stage 1 question types and their corresponding templates.}
  \label{tab:question_templates}
\end{table}

\noindent
\textbf{QA Generation}. Questions need to accurately reflect the four defined intents, and answers must be exact. We created templates for each question–answer pair in Table~\ref{tab:question_templates}. To make them read more fluently, we used a large language model to rephrase both questions and answers.

\noindent
\textbf{VP Rendering}. To generate diverse VPs, we developed a modular rendering framework. It supports many VP shapes by using the factory pattern in the core visualization module and the proxy pattern at the interface layer. Our implementation defines 8 independent VP‐shape classes; new shapes can be added without altering existing code. We integrate this into our data pipeline via a unified drawing interface with customizable attributes, improving readability and reducing development effort.

\newcolumntype{L}[1]{>{\raggedright\arraybackslash}p{#1}} 
\begin{table}[ht]
  \centering
  \small
  \begin{tabular}{@{}L{1cm} L{6.5cm}@{}}
    \toprule
    \textbf{Scenario} & \textbf{Question Template} \\
    \midrule
    FER   & What emotion is the person showing in the area highlighted by \texttt{<color>} \texttt{<VP>}? \\
    SVR   & What does the red oval highlight in this \texttt{<color>} \texttt{<VP>}? \\
    MIA   & A \texttt{<Medical Image Type>} of \texttt{<Human Organ>} is provided, with the region of interest marked by \texttt{<color>} \texttt{<VP>}. Please assess the pathological characteristics of this region. \\
    3DOR  & Given a top‑down and a front view (from south to north) of a room, identify the object marked by the \texttt{<color>} \texttt{<VP>} and specify its position. \\
    SGG   & What is the object that is on the thing highlighted by the \texttt{<color>} \texttt{<VP>}? \\
    GUIER & Can you recognize the graphical user interface (GUI) element highlighted by the \texttt{<color>} \texttt{<VP>}? \\
    \bottomrule
  \end{tabular}
  \caption{Stage 2 downstream‑task question templates: Facial Emotion
           Recognition (FER), Street‑view Recognition (SVR), Medical Image
           Analysis (MIA), 3D Object Recognition (3DOR), Scene-Graph Generation (SGG), and GUI Element Recognition (GUIER).}
  \label{tab:app_question_templates}
\end{table}

\subsection{Stage 2 Data Curation}
\noindent
\textbf{Question Definition}. Stage 2 examines how different VPs affect downstream tasks. We selected six common tasks that also involve reference information:
\begin{enumerate}
    \item Medical Image Analysis (MIA),
    \item Indoor 3D Object Recognition,
    \item Facial Expression Recognition,
    \item Street-Scene Recognition,
    \item GUI Element Recognition,
    \item Scene‐Graph Generation (SGG).
\end{enumerate}
For each task, we retain its original objective. For example, in medical‐image analysis, the question becomes: “Identify the region highlighted by the VP and diagnose the possible condition.”

\textbf{Data Collection}. Each co-author was assigned one downstream task and located the corresponding training or validation dataset by: (i) original publication datasets, those introduced by the task’s authors; (ii) public repositories, any relevant dataset on HuggingFace\footnote{\url{https://huggingface.co/}}, Papers with Code\footnote{\url{https://paperswithcode.com/}}, etc. Each dataset then review by three independent examiners (excluding its collector) to ensure:
\begin{enumerate}
    \item license availability, usable under our intended license,
    \item VP applicability, supports overlaying visual prompts,
    \item annotation quality, those clear, accurate labels to avoid ambiguous or incorrect QAs.
\end{enumerate}
Approved datasets were converted into the metadata schema defined in Table~\ref{tab:meta_example}.

\textbf{QA Generation}. To ensure consistency and clarity, we designed standardized question templates tailored to each task. For example, in the facial emotion recognition task, the template takes the form:
\begin{quote}
  \textit{What emotion is the person showing in the area highlighted by \texttt{<color>} \texttt{<VP>}?}
\end{quote}
To generate natural distractors, we randomly selected three incorrect emotion categories from the dataset and prompted the LLM to produce three fluent but misleading alternatives.

For the MIA (Medical Image Analysis) task, the template follows a similar pattern:
\begin{quote}
  \textit{A \texttt{<Medical Image Type>} of \texttt{<Human Organ>} is provided, with the region of interest marked by \texttt{<color>} \texttt{<VP>}. Please assess the pathological characteristics of this region.}
\end{quote}
Here, distractors were generated by providing the LLM with incorrect tissue types and unrelated disease causes to construct plausible but incorrect options.

Likewise, in the SVR (Street-View Recognition) task, the question format is:
\begin{quote}
  \textit{What does the red oval highlight in this \texttt{<color>} \texttt{<VP>}?}
\end{quote}
The LLM was guided using unrelated traffic components as cues, ensuring the distractors appeared contextually relevant but remained distinguishable from the correct answer.

\noindent
\textbf{VP Rendering}. Thanks to the Stage 1 framework, we simply reused the same codebase to render the visual prompts needed for Stage 2.

\begin{table}[htbp]
  \centering
  \small
  \resizebox{\linewidth}{!}{%
  \begin{tabular}{lcc}
    \toprule
    \textbf{Model} & \textbf{Avg. VP Accuracy} & \textbf{Avg. Downstream Accuracy} \\
    \midrule
    Human Baseline & 90.03 & -- \\
    \midrule
    GPT-4o (proprietary) & 68.80 & 72.56 \\
    Qwen-VL-Max (proprietary) & 82.63 & 75.61 \\
    \midrule
    Qwen2.5-VL-3B & 73.41 & 65.76 \\
    Qwen2.5-VL-7B & 81.29 & 69.08 \\
    Qwen2.5-VL-32B & 83.22 & 70.93 \\
    Qwen2.5-VL-72B & 82.80 & 75.55 \\
    \midrule
    InternVL3-1B & 71.46 & 55.98 \\
    InternVL3-8B & 84.11 & 72.89 \\
    InternVL3-38B & 86.89 & 77.17 \\
    InternVL3-78B & \textbf{87.97} & \textbf{79.05} \\
    \midrule
    Molmo-72B & 85.61 & 76.18 \\
    NVLM-D-72B & 85.39 & 74.67 \\
    \bottomrule
  \end{tabular}}%
  \caption{Simplified summary of model scale versus VP perception (Stage 1) and downstream task accuracy (Stage 2).}
  \label{tab:vp_scale_summary}
\end{table}

\begin{table*}[h]
  \centering
  \resizebox{0.99\textwidth}{!}{%
  \begin{tabular}{l *{12}{c} c c}
    \toprule
    \multirow{2}{*}{Model} &
    \multicolumn{2}{c}{Emotic} &
    \multicolumn{2}{c}{Mapillary Vistas} &
    \multicolumn{2}{c}{MIA} &
    \multicolumn{2}{c}{SD‑100} &
    \multicolumn{2}{c}{SGG} &
    \multicolumn{2}{c}{See‑Click} &
    Avg.\ Vis. ($\Delta$) & Avg.\ Txt. \\
    \cmidrule(lr){2-3}\cmidrule(lr){4-5}\cmidrule(lr){6-7}\cmidrule(lr){8-9}\cmidrule(lr){10-11}\cmidrule(lr){12-13}
    &
      Vis. & Txt. &
      Vis. & Txt. &
      Vis. & Txt. &
      Vis. & Txt. &
      Vis. & Txt. &
      Vis. & Txt. &
           &      \\ \midrule
    CogVLM2‑LLama3‑Chat‑19B & 70.71 & 78.12 & 33.49 & 23.92 &  5.40 &  5.60 & 31.33 & 13.33 & 83.30 & 83.30 & 75.31 & 43.50 & 49.92 (+8.86) & 41.30 \\
    DeepSeek‑VL2            & 20.90 & 67.76 & 24.88 & 24.40 & 20.90 & 20.60 & 45.67 & 19.33 & 83.30 & 84.37 & 67.50 & 68.00 & 43.86 (-3.55) & 47.41 \\
    GLM‑4V‑9B               & 64.94 & 69.88 & 53.59 & 30.62 & 18.19 & 18.70 & 59.33 & 40.00 & 86.32 & 86.59 & 96.25 & 95.00 & 63.10 (+6.31) & 56.80 \\
    Llama‑3.2‑90B‑Vision‑Instruct & 65.65 & 69.65 & 51.20 & 44.98 & 59.50 & 62.70 & 58.00 & 56.00 & 88.99 & 92.18 & 98.50 & 97.50 & 70.31 (-0.20) & 70.50 \\
    LLaVA‑OneVision‑Qwen2‑7B‑OV‑HF & 64.39 & 74.29 & 47.37 & 33.49 & 16.72 & 17.72 & 63.67 & 63.33 & 92.01 & 90.76 & 92.01 & 94.25 & 62.70 (+0.39) & 62.31 \\
    LLaVA‑v1.6‑34B‑HF       & 64.05 & 64.47 & 48.33 & 43.06 & 37.60 & 35.30 & 52.33 & 55.00 & 91.30 & 89.52 & 96.30 & 94.75 & 64.99 (+1.30) & 63.68 \\
    MiniCPM‑V‑2\_6          & 67.06 & 72.71 & 26.32 & 35.41 & 15.30 & 13.70 & 38.00 & 65.33 & 89.08 & 89.88 & 92.75 & 96.75 & 54.75 (-7.55) & 62.30 \\
    Molmo‑72B‑0924          & 67.38 & 61.88 & 60.77 & 52.15 & 62.80 & 62.90 & 78.00 & 67.33 & 91.83 & 86.59 & 96.30 & 89.75 & 76.18 (+6.08) & 70.10 \\
    NVLM‑D‑72B             & 72.71 & 73.18 & 49.76 & 41.15 & 58.10 & 60.90 & 77.00 & 75.67 & 93.43 & 92.18 & 97.00 & 96.75 & 74.67 (+1.36) & 73.31 \\
    Ovis2‑34B              & 73.10 & 74.12 & 50.72 & 60.77 & 37.30 & 50.20 & 83.33 & 79.00 & 94.58 & 92.36 & 99.69 & 99.25 & 73.12 (-2.83) & 75.95 \\
    Qwen2.5‑VL‑72B         & 71.43 & 71.53 & 61.72 & 59.81 & 43.80 & 39.20 & 83.33 & 85.67 & 95.47 & 94.14 & 97.53 & 98.00 & 75.55 (+0.82) & 74.73 \\
    \bottomrule
  \end{tabular}}%
  \caption{Comparison of model performance with VP and text-based spatial prompt (TP) baseline on Stage 2 evaluation. Average VP accuracy reports absolute accuracy followed by the signed changed $\Delta$ relative to its baseline.}
  \label{tab:prompt_method_comparison}
\end{table*}

\section{More Experiments Results and Analysis}

\begin{table*}[h]
  \centering
  \resizebox{0.99\textwidth}{!}{%
  \begin{tabular}{l*{8}{c}*{4}{c}c}
    \toprule
    \multirow{2}{*}{Model} &
      \multicolumn{8}{c}{\textbf{Visual Prompt Types}} &
      \multicolumn{4}{c}{\textbf{Problem Types}} &
      \multirow{2}{*}{Avg.} \\
    \cmidrule(lr){2-9}\cmidrule(lr){10-13}
      & Tag & Arrow & BBox & Contour & Mask & Oval & Point & Scribble &
      Enumeration & Existence & Rough-Loc. & VP-Ref. \\
    \midrule
\multicolumn{14}{l}{\textit{Human Baseline}} \\
Human Reviewers & 89.00 & 92.62 & 97.29 & 87.68 & 85.28 & 94.87 & 90.68 & 82.80 & 90.73 & 94.36 & 97.68 & 84.26 & 90.03 \\
\midrule
\multicolumn{14}{l}{\textit{Proprietary Models}} \\
GPT-4o & 69.95 & 70.27 & 74.18 & 77.18 & 65.32 & 79.77 & 49.28 & 64.45 & 60.44 & 87.03 & 57.83 & 67.74 & 68.80 \\
Doubao-Seed-1.6 & 86.21 & 79.60 & 93.60 & 72.43 & 71.19 & 50.17 & 47.49 & 62.30 & 62.47 & 64.04 & 95.34 & 80.53 & 70.37 \\
Qwen-VL-Max & 92.27 & 82.84 & 93.10 & 88.75 & 67.51 & 92.84 & 69.31 & 74.45 & 81.80 & 88.11 & 92.13 & 76.82 & 82.63 \\
\midrule
\multicolumn{14}{l}{\textit{Pre-trained Models}} \\
Qwen2.5-VL-72B-Instruct & 92.26 & 82.57 & 92.88 & 88.62 & 68.68 & 92.83 & 69.58 & 74.77 & 81.84 & 88.26 & 92.01 & 75.79 & 82.80 \\
NVLM-D-72B & 91.28 & 88.23 & 91.60 & 90.22 & 76.02 & 93.35 & 77.00 & 75.41 & 83.82 & 93.99 & 92.06 & 76.50 & 85.39 \\
Molmo-72B-0924 & 90.71 & 85.31 & 92.92 & 89.61 & 79.05 & 93.18 & 76.57 & 77.56 & 87.18 & 97.24 & 92.46 & 70.08 & 85.61 \\
InternVL3-78B & 93.87 & 85.77 & 94.25 & 91.56 & 80.01 & 95.81 & 81.59 & 80.89 & 88.59 & 92.93 & 95.24 & 79.65 & 87.97 \\
    \bottomrule
  \end{tabular}}%
  \caption{Performance on VP‐shape and question‐type subtasks, sorted by overall accuracy.}
  \label{tab:model_comparison}
\end{table*}

All experiments were conducted on a cluster of 8$\times$H800 GPUs (80\,GB memory each). As no training was performed, the computational cost is limited to model inference. During inference, the decoding was restricted to the top-1 response for all evaluations to eliminate randomness.

\textbf{Human performance baseline compared with model results.} To establish a human performance baseline, participants unfamiliar with this work individually completed the Stage~1 benchmark. For efficiency, 1,000 questions were randomly sampled from the original dataset, and each participant took the test independently without any prior briefing or in-process discussion.

Human participants achieved an average accuracy of 90.03\% across all tasks, whereas the best-performing model, InternVL3-78B, reached 87.97\%. This indicates that while current models exhibit near-human VP perception, there remains room for improvement. In certain VP types, however, the model surpassed human performance; for example, InternVL3‑78B achieved 93.87\% on \textbf{Tag}, exceeding the human score of 89.00\%. Nevertheless, model stability lagged behind that of humans. In \textbf{Point} perception, for instance, human accuracy was 90.68\%, whereas model performance ranged from 49.28\% to 81.59\%.

\textbf{Open-source models have already overtaken proprietary models in VP perception.} In Table~\ref{tab:model_comparison}, most tested open-source models achieve an average accuracy of around 80\%, with InternVL3-78B reaching 88.62\%. By contrast, proprietary models score 68.93\% for GPT-4o, 72.13\% for Doubao-Seed-1.6, and 83.7\% for Qwen-VL-Max. Surprisingly, GPT-4o’s spatial perception of VPs lags considerably, at only 57.83\% accuracy. In our Stage~2 evaluation on downstream tasks, NVLM-D-72B, Qwen-2.5-VL-72B, InternVL3-78B, and Molmo-72B perform very competitively against proprietary models. Notably, Molmo-72B outperforms GPT-4o on MIA, street-view recognition, 3D object recognition, and SGG tasks. Taken together, these findings indicate that open-source MLLMs are not only closing the gap but have already surpassed proprietary counterparts in VP perception and several downstream domains.

\textbf{Model scale strongly correlates with VP perception accuracy.} As shown in Table~\ref{tab:vp_scale_summary}, both the Qwen2.5-VL and InternVL3 families exhibit consistent performance gains with increasing parameter size, with InternVL3-78B achieving the highest Stage~1 accuracy 87.97\%. Larger models demonstrate particular advantages on irregular VP types such as Mask and Point, where smaller models show pronounced drops, suggesting that fine-grained regional perception benefits from higher-capacity vision–language representations. For Stage~2 downstream tasks: models above 30B parameters, including InternVL3-38B, InternVL3-78B, Molmo-72B, and NVLM-D-72B, consistently outperform mid- and small-scale counterparts across MIA, 3D object recognition, and street-view recognition. These results indicate that scaling improves both VP detection and the integration of VP cues into task-specific reasoning, narrowing the gap to human-level performance and enabling robust transfer to diverse application domains.

\section{Prompts}
Accurate prompt design is critical for consistent data generation, annotation refinement, and model evaluation in VP-Bench. This section details the prompt templates used across three stages of the pipeline. First, Table~\ref{tab:prompt_distr} presents the multi-choice question generation prompt, which instructs the LLM to produce plausible but incorrect distractors to enhance dataset diversity. Second, Table~\ref{tab:prompt_expand} shows the annotation expansion prompt, which rephrases and reorganizes textual descriptions while retaining their core semantics to improve linguistic variety. Finally, Tables~\ref{tab:prompt_bench} and \ref{tab:prompt_examples_combined} describe the benchmarking prompts that standardize model evaluation using a consistent question–answering format, with or without explicit VP descriptions, enabling controlled experiments on VP perception.


\begin{table}[h]
\centering
\small
\begin{tabular}{p{\columnwidth}}
\hline
\begin{minipage}[t]{\columnwidth}
\scriptsize
\textless SYSTEM PROMPT\textgreater

You are an AI assistant specialized in generating plausible distractor options for multiple-choice visual questions. Your primary task is to create distractor options that are incorrect but plausible, aiming to challenge the user while ensuring it remains distinguishable from the correct answer.\\

\textless USER PROMPT\textgreater

For the question '\textless question\textgreater' and its correct answer '\textless answer\textgreater', generate three plausible but incorrect distractor options. These options should be similar to the correct answer to create confusion, yet still distinct enough to be clearly wrong.\\

Your output should be formatted as follows in JSON:
\verb|```json|\\[2pt]
\verb|{|\\[2pt]
\verb|    "distractor_options": [|\\[2pt]
\verb|        "Distractor option 1",|\\[2pt]
\verb|        "Distractor option 2",|\\[2pt]
\verb|        "Distractor option 3"|\\[2pt]
\verb|    ]|\\[2pt]
\verb|}|\\[2pt]
\verb|```|\\[2pt]
\end{minipage} \\ \hline
\end{tabular}
\caption{Prompt template used to instruct the LLM to generate plausible but incorrect distractor options for multiple-choice visual questions. The system prompt defines the task, while the user prompt supplies the question and the correct answer.}
\label{tab:prompt_distr}
\end{table}


\begin{table}[h]
    \centering
    \scriptsize
    \begin{tabular}{p{\columnwidth}}
        \midrule
        The following paragraph should be rewritten while retaining the essential information. Different expressions should be used, and the paragraph may be reorganized if necessary. The paragraph should not be altered merely by converting the passive voice to active voice or vice versa.\\
        \midrule
    \end{tabular}
    \caption{Instruction template for rewriting a paragraph while preserving its essential information. The guidance emphasizes using different expressions and allows reorganization of the content rather than simply switching between passive and active voice.}
    \label{tab:prompt_expand}
\end{table}


\begin{table}[h]
\centering
\small
\begin{tabular}{p{\columnwidth}}
\hline
\begin{minipage}[t]{\columnwidth}
\scriptsize
Context: \{VP description\}\\[1ex]
\{QUESTION\}\\[1ex]
A. \{OPTION\_A\}\\[0.5ex]
B. \{OPTION\_B\}\\[0.5ex]
C. \{OPTION\_C\}\\[0.5ex]
D. \{OPTION\_D\}\\[1ex]
Answer with the option’s letter from the given choices directly.
\end{minipage}\\
\hline
\end{tabular}
\caption{General instruction prompt template used to generate multiple-choice questions for VP perception tasks. The \{VP description\} field is optionally included depending on the experimental setting.}
\label{tab:prompt_bench}
\end{table}

\begin{table}[h]
\centering
\small
\begin{tabular}{p{0.48\columnwidth}p{0.48\columnwidth}}
\toprule
\multicolumn{1}{c}{\textit{Example with VP description}} & 
\multicolumn{1}{c}{\textit{Example without VP description}} \\
\midrule
\begin{minipage}[t]{0.48\columnwidth}
\scriptsize
Context: An arrow is a symbol that points to a target, which may be in red, blue, green, or black.\\[1ex]
Where is the red arrow located in the image?\\[1ex]
A. The red arrow is located in the bottom-left.\\[0.5ex]
B. The red arrow is located in the bottom-right.\\[0.5ex]
C. The red arrow is located in the top-right.\\[0.5ex]
D. The red arrow is located in the top-left.\\[1ex]
Answer with the option’s letter from the given choices directly.
\end{minipage} & 
\begin{minipage}[t]{0.48\columnwidth}
\scriptsize
Where is the red arrow located in the image?\\[1ex]
A. The red arrow is located in the bottom-left.\\[0.5ex]
B. The red arrow is located in the bottom-right.\\[0.5ex]
C. The red arrow is located in the top-right.\\[0.5ex]
D. The red arrow is located in the top-left.\\[1ex]
Answer with the option’s letter from the given choices directly.
\end{minipage}\\
\bottomrule
\end{tabular}
\caption{Comparison of instruction arrangements with and without a VP description. Left: Guiding the model to interpret the visual prompt via the VP description; Right: Testing the model’s ability to infer VP meaning from visual cues alone.}
\label{tab:prompt_examples_combined}
\end{table}

\begin{sidewaystable*}
\centering
\resizebox{\textwidth}{!}{%
\begin{tabular}{|l|l|l|l|l|l|}
\hline
\textbf{Visual Prompt Shape} & \textbf{Attribute 1} & \textbf{Attribute 2} & \textbf{Attribute 3} & \textbf{Attribute 4} & \textbf{Attribute 5} \\ \hline
\multirow{4}{*}{Tag} 
  & Red Color Tag & Blue Color Tag & Green Color Tag & Black Color Tag & Background Contrastive Color Tag \\ \cline{2-6}
  & Alphabet    & Digit         &               &               &                          \\ \cline{2-6}
  & Square      & Round         &               &               &                          \\ \cline{2-6}
  & Small Front & Medium Front  & Big Front     &               &                          \\ \hline
\multirow{3}{*}{Bounding Box} 
  & Red Color Line & Blue Color Line & Green Color Line & Black Color Line & Background Contrastive Color Line \\ \cline{2-6}
  & Thin Line    & Normal Width Line & Thick Line  &               &                          \\ \cline{2-6}
  & Vertex No Shape & Vertex Square Shape & Vertex Round Shape &       &                          \\ \hline
\multirow{3}{*}{Arrow} 
  & Red Color Line & Blue Color Line & Green Color Line & Black Color Line & Background Contrastive Color Line \\ \cline{2-6}
  & Thin Line    & Normal Width Line & Thick Line  &               &                          \\ \cline{2-6}
  & Arrow Bold Style & Arrow Thin Style &           &               &                          \\ \hline
\multirow{3}{*}{Mask} 
  & Red Color Line (Filled) & Blue Color Line (Filled) & Green Color Line (Filled) & Black Color Line (Filled) & Background Contrastive Color Line (Filled) \\ \cline{2-6}
  & Thin Line    & Normal Width Line & Thick Line  &               &                          \\ \cline{2-6}
  & Contour with Filled & Contour Only &           &               &                          \\ \hline
\multirow{3}{*}{Contour} 
  & Red Color Line & Blue Color Line & Green Color Line & Black Color Line & Background Contrastive Color Line \\ \cline{2-6}
  & Thin Line    & Normal Width Line & Thick Line  &               &                          \\ \cline{2-6}
  & Strict Contour & Loosen Contour &           &               &                          \\ \hline
\multirow{2}{*}{Oval} 
  & Red Color Line & Blue Color Line & Green Color Line & Black Color Line & Background Contrastive Color Line \\ \cline{2-6}
  & Thin Line    & Normal Width Line & Thick Line  &               &                          \\ \hline
\multirow{3}{*}{Point} 
  & Red Color Dot & Blue Color Dot & Green Color Dot & Black Color Dot & Background Contrastive Color Dot \\ \cline{2-6}
  & Small Size  & Medium Size   & Big Size     &               &                          \\ \cline{2-6}
  & Round Shape & Square Shape  &              &               &                          \\ \hline
\multirow{2}{*}{Scribble} 
  & Red Color Line & Blue Color Line & Green Color Line & Black Color Line & Background Contrastive Color Line \\ \cline{2-6}
  & Thin Line    & Normal Width Line & Thick Line  &               &                          \\ \hline
\end{tabular}%
}
\caption{Visual prompt shapes and their attributes.}
\label{tab:vp_shapes_attributes}
\end{sidewaystable*}

\section{Visual Prompt Properties}
Visual prompts are visual cues embedded within images that help guide a model's attention and interpretation. They come in a variety of shapes, such as tags, bounding boxes, arrows, masks, contours, ovals, points, and scribbles, each designed with distinct properties. These properties include color variations, line styles, thicknesses, and additional stylistic features that can be adjusted to enhance contrast and clarity. Table~\ref{tab:vp_shapes_attributes} provides a detailed overview of these visual prompt shapes along with a set of attributes for each, illustrating the customizable nature of these prompts and how they can be tailored for specific applications.

\section{Qualitative Result}
\begin{figure*}[h]
    \centering
    \includegraphics[width=1\linewidth]{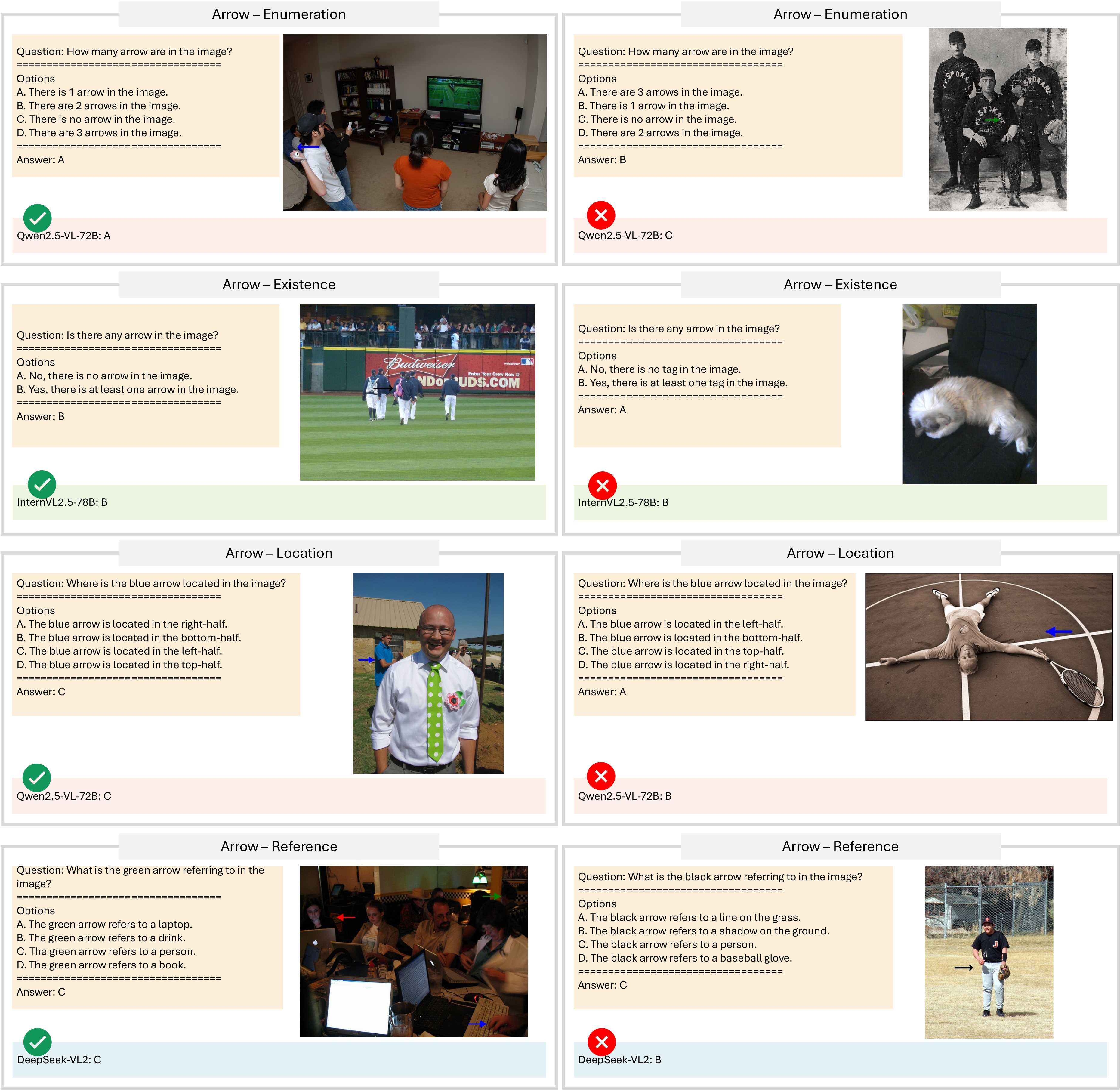}
    \caption{Qualitative results of arrow shape in Stage 1 evaluation.}
    \label{fig:stage-1-arrow}
\end{figure*}

\begin{figure*}[h]
    \centering
    \includegraphics[width=1\linewidth]{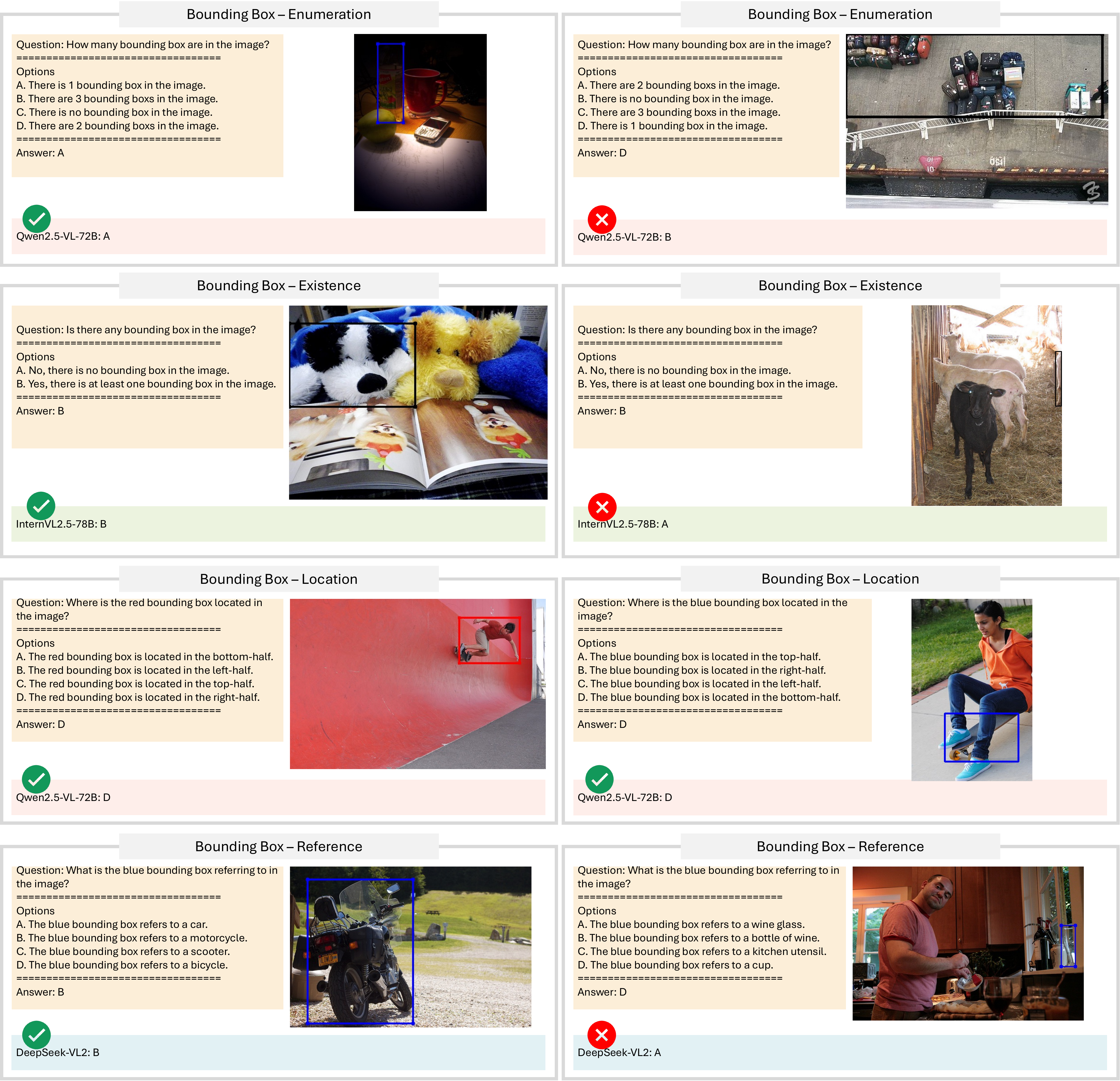}
    \caption{Qualitative results of bounding box shape in Stage 1 evaluation.}
    \label{fig:stage-1-bbox}
\end{figure*}

\begin{figure*}[h]
    \centering
    \includegraphics[width=1\linewidth]{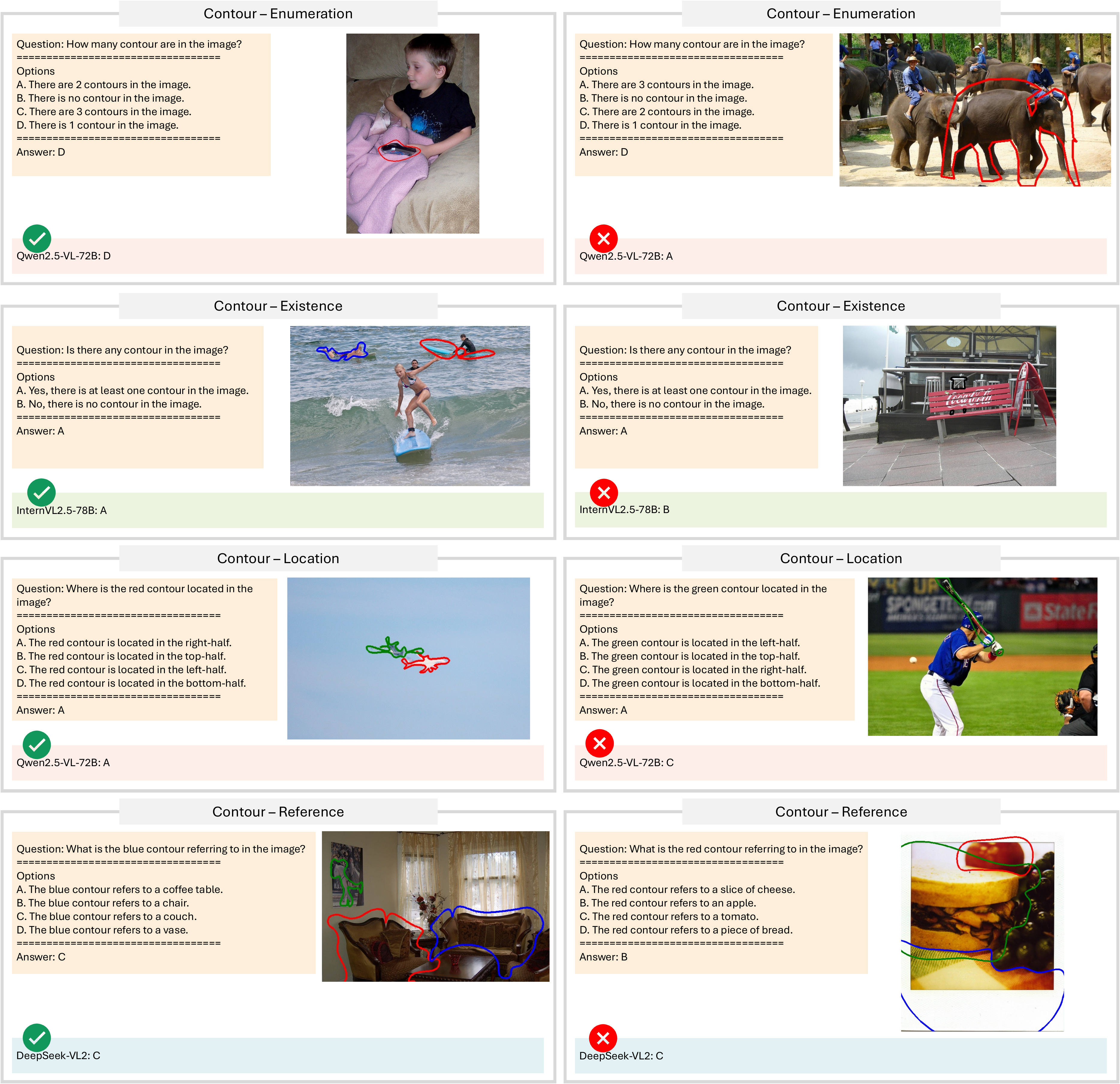}
    \caption{Qualitative results of contour shape in Stage 1 evaluation.}
    \label{fig:stage-1-contour}
\end{figure*}

\begin{figure*}[h]
    \centering
    \includegraphics[width=1\linewidth]{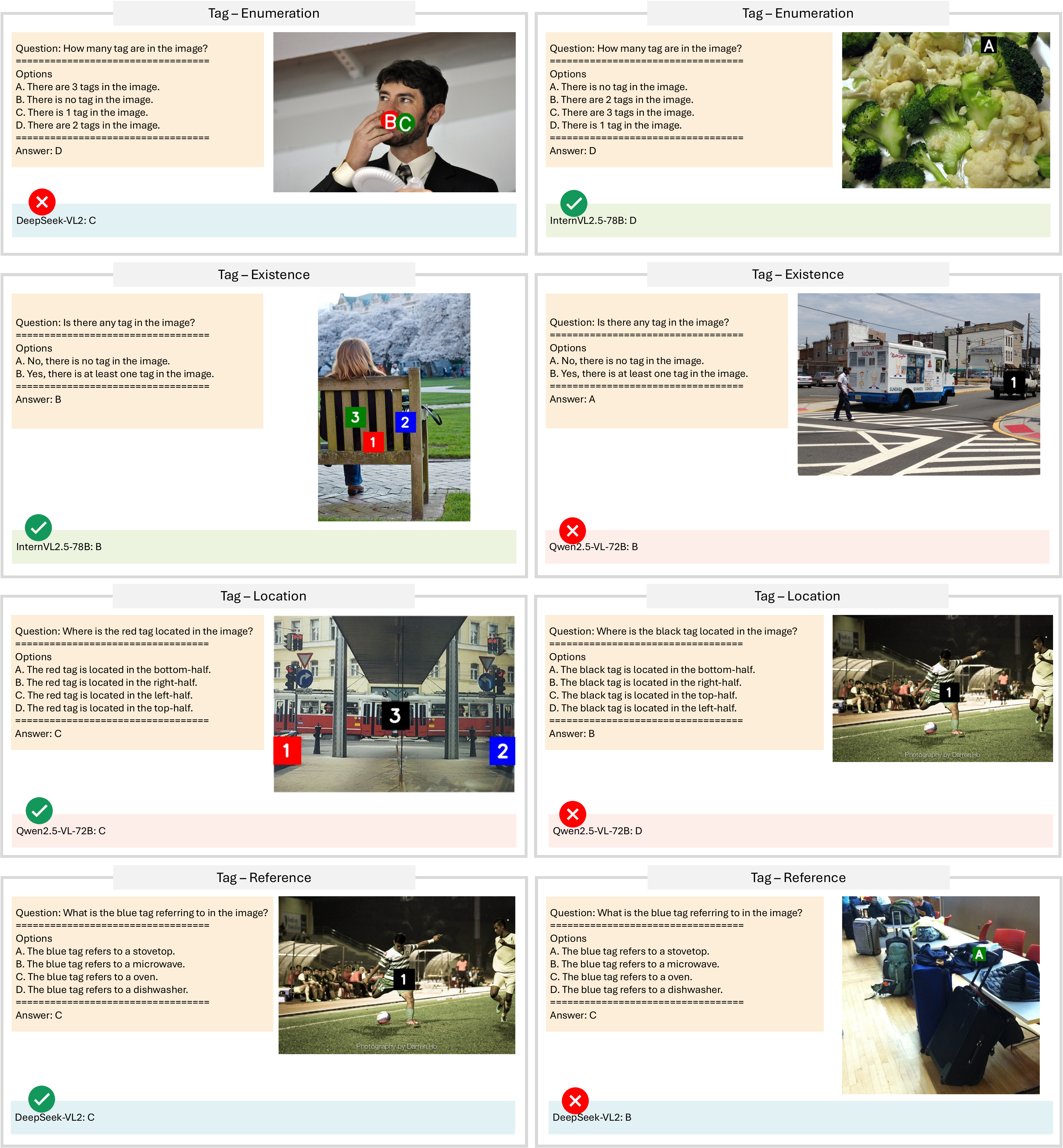}
    \caption{Qualitative results of tag shape in Stage 1 evaluation.}
    \label{fig:stage-1-tag}
\end{figure*}

\begin{figure*}[h]
    \centering
    \includegraphics[width=1\linewidth]{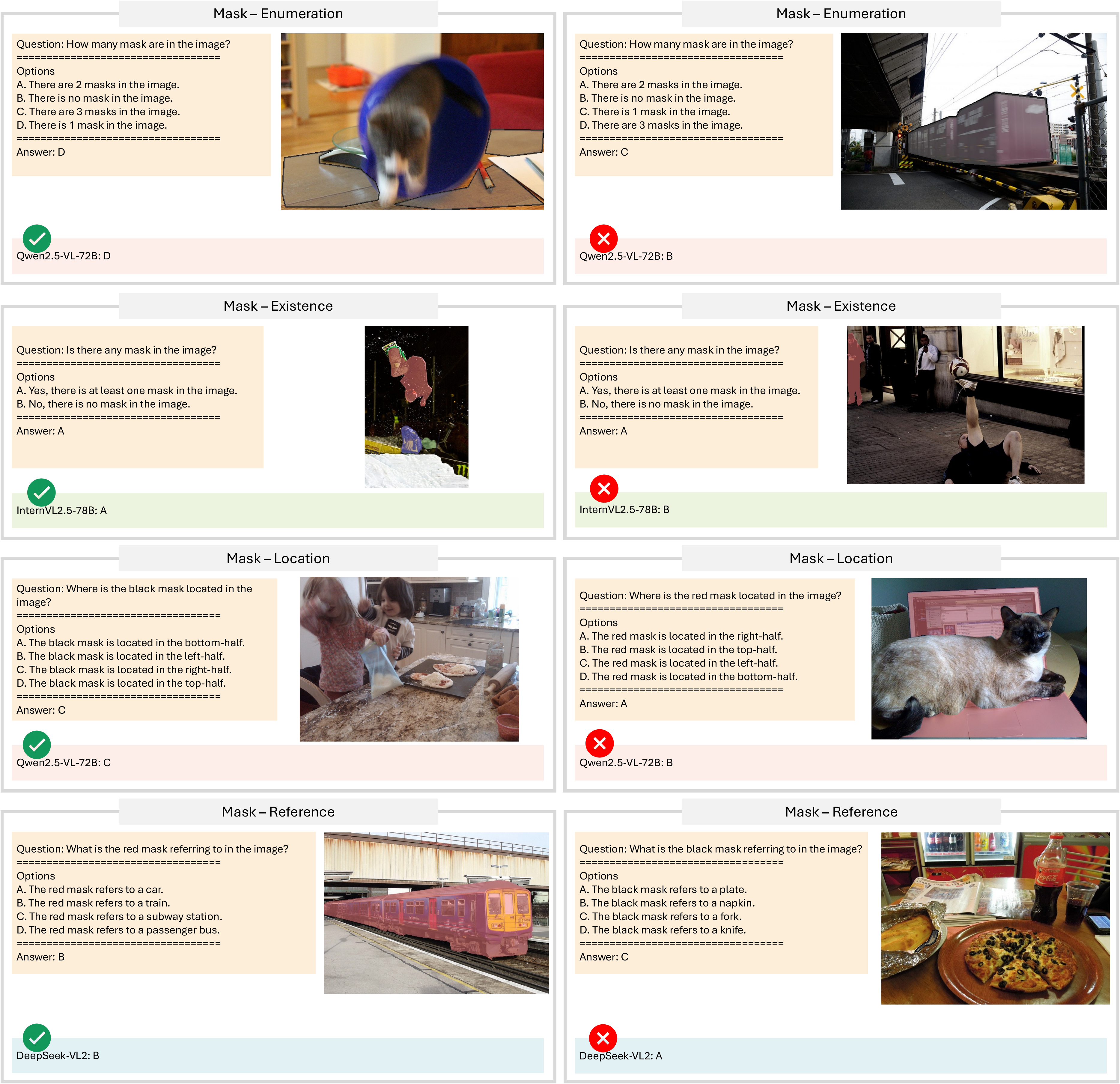}
    \caption{Qualitative results of mask shape in Stage 1 evaluation.}
    \label{fig:stage-1-mask}
\end{figure*}

\begin{figure*}[h]
    \centering
    \includegraphics[width=1\linewidth]{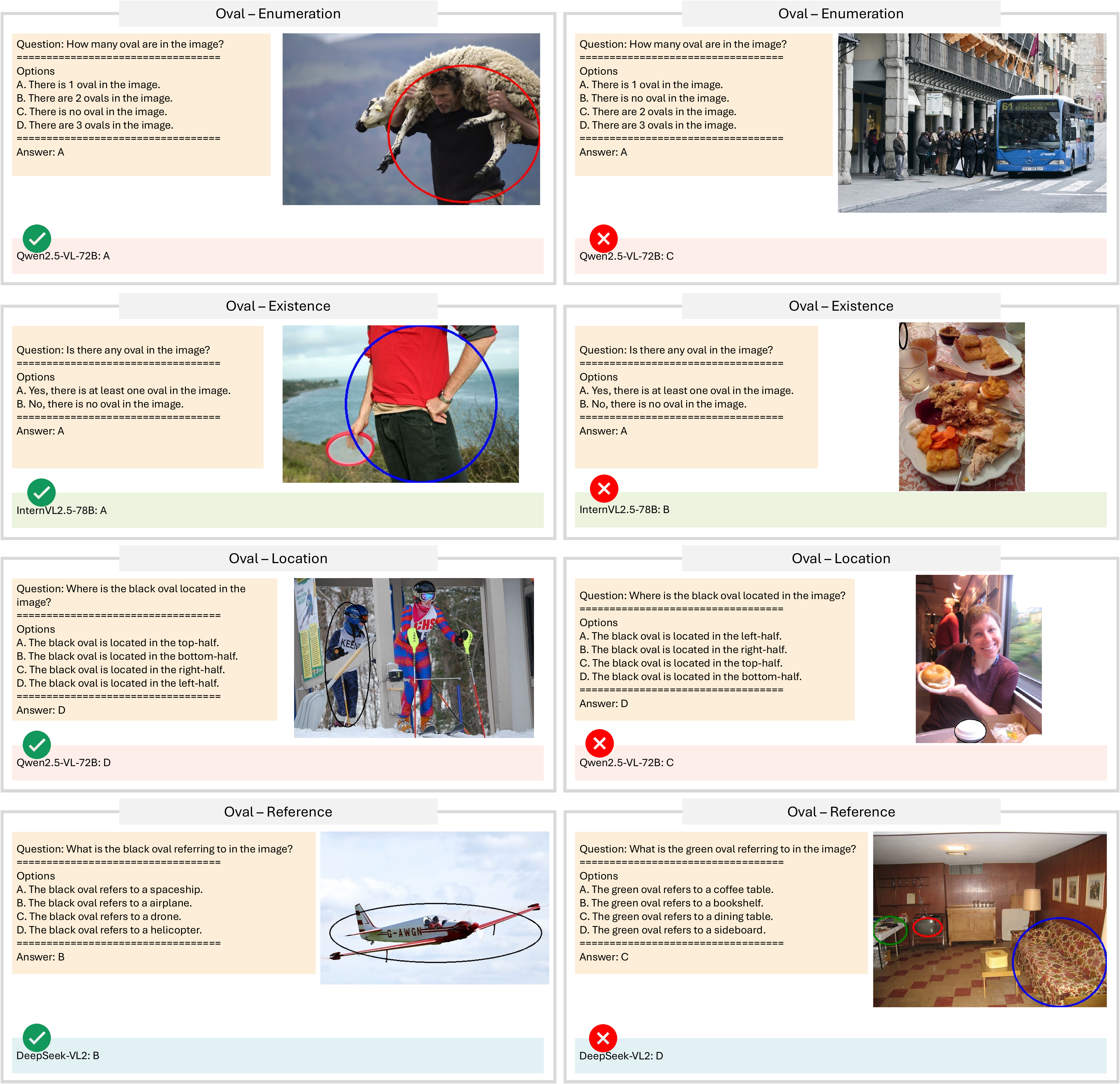}
    \caption{Qualitative results of oval shape in Stage 1 evaluation.}
    \label{fig:stage-1-oval}
\end{figure*}

\begin{figure*}[h]
    \centering
    \includegraphics[width=1\linewidth]{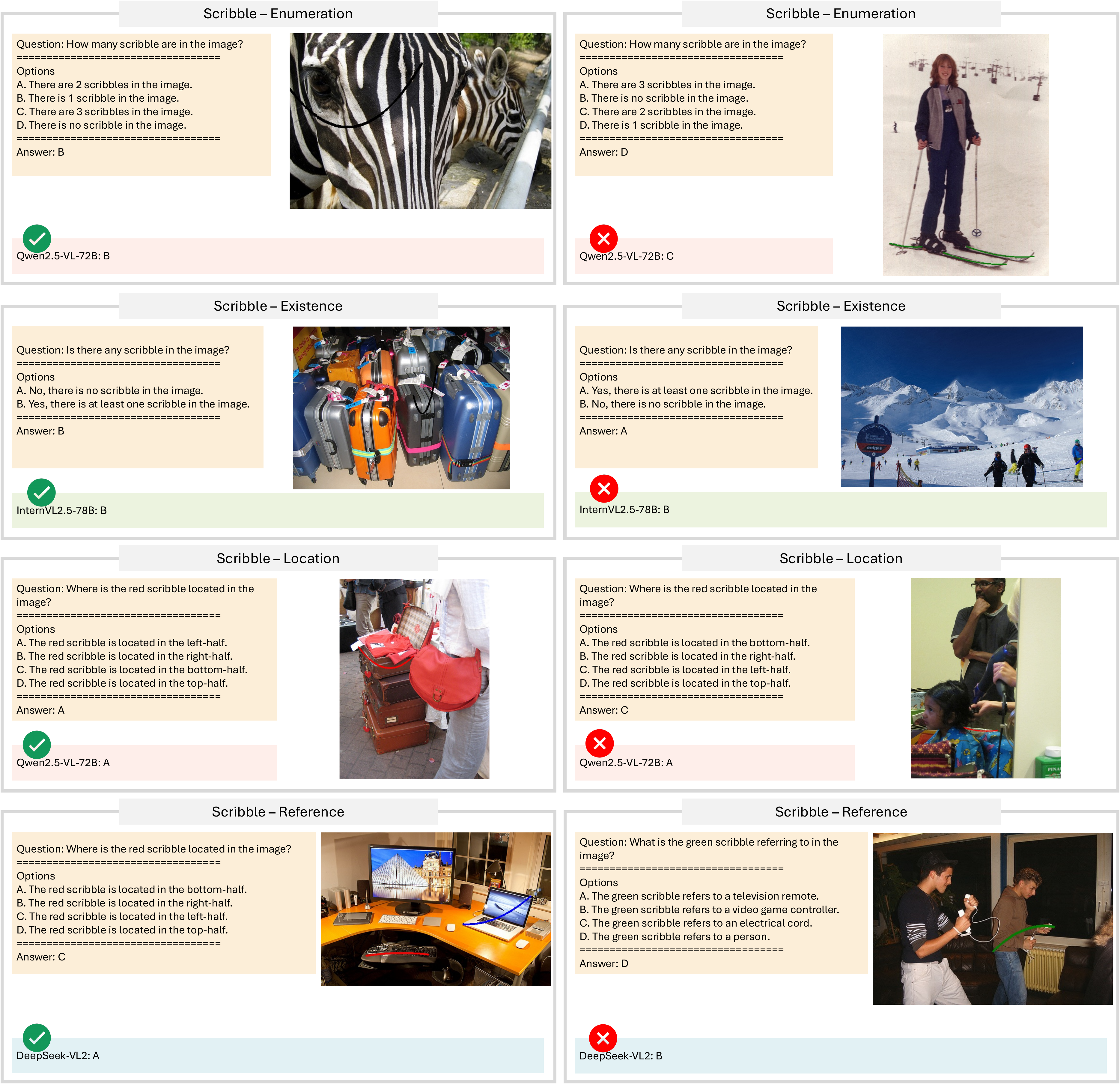}
    \caption{Qualitative results of scribble shape in Stage 1 evaluation.}
    \label{fig:stage-1-scribble}
\end{figure*}

\begin{figure*}[h]
    \centering
    \includegraphics[width=1\linewidth]{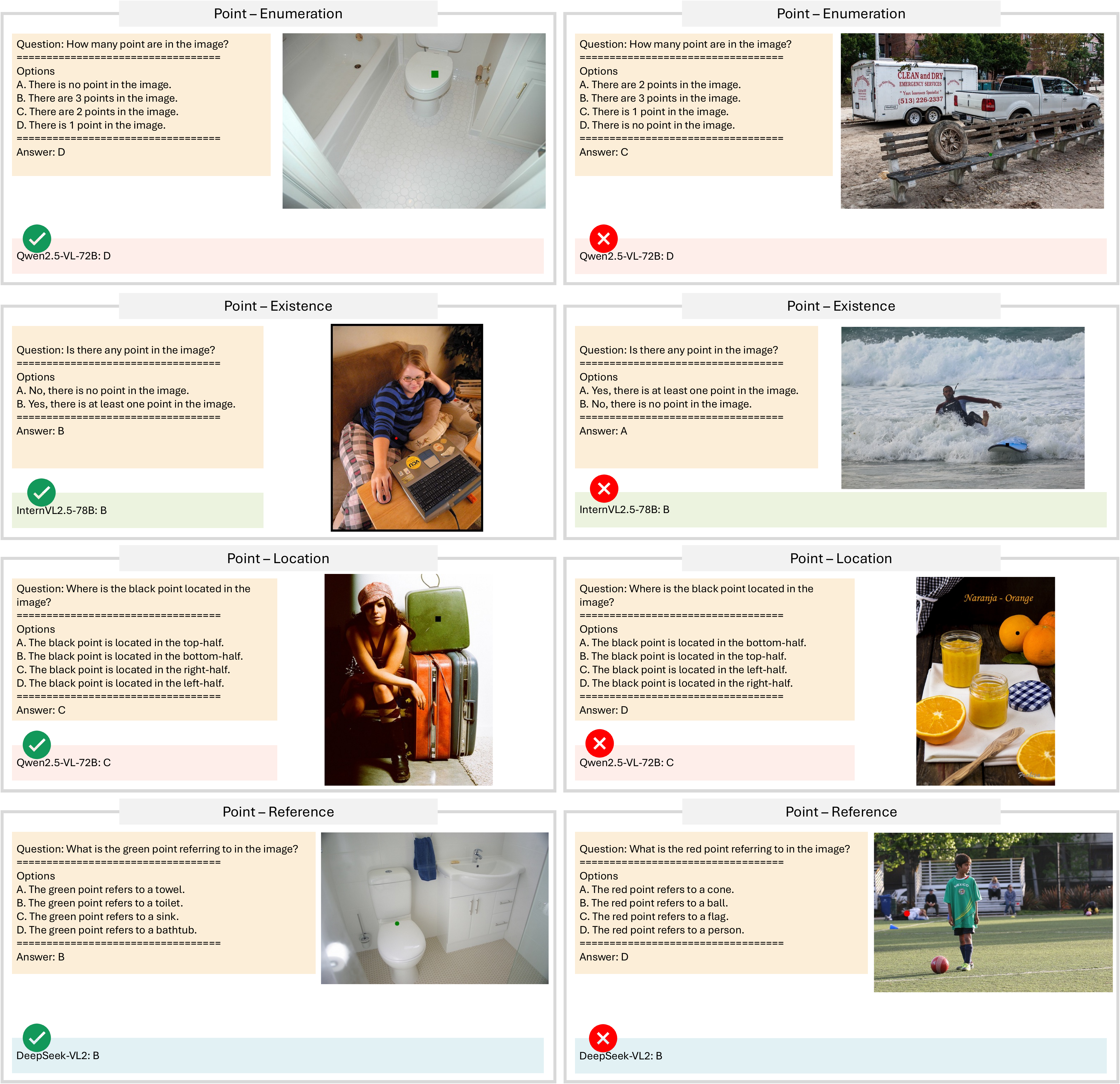}
    \caption{Qualitative results of point shape in Stage 1 evaluation.}
    \label{fig:stage-1-point}
\end{figure*}

\begin{figure*}[h]
    \centering
    \includegraphics[width=1\linewidth]{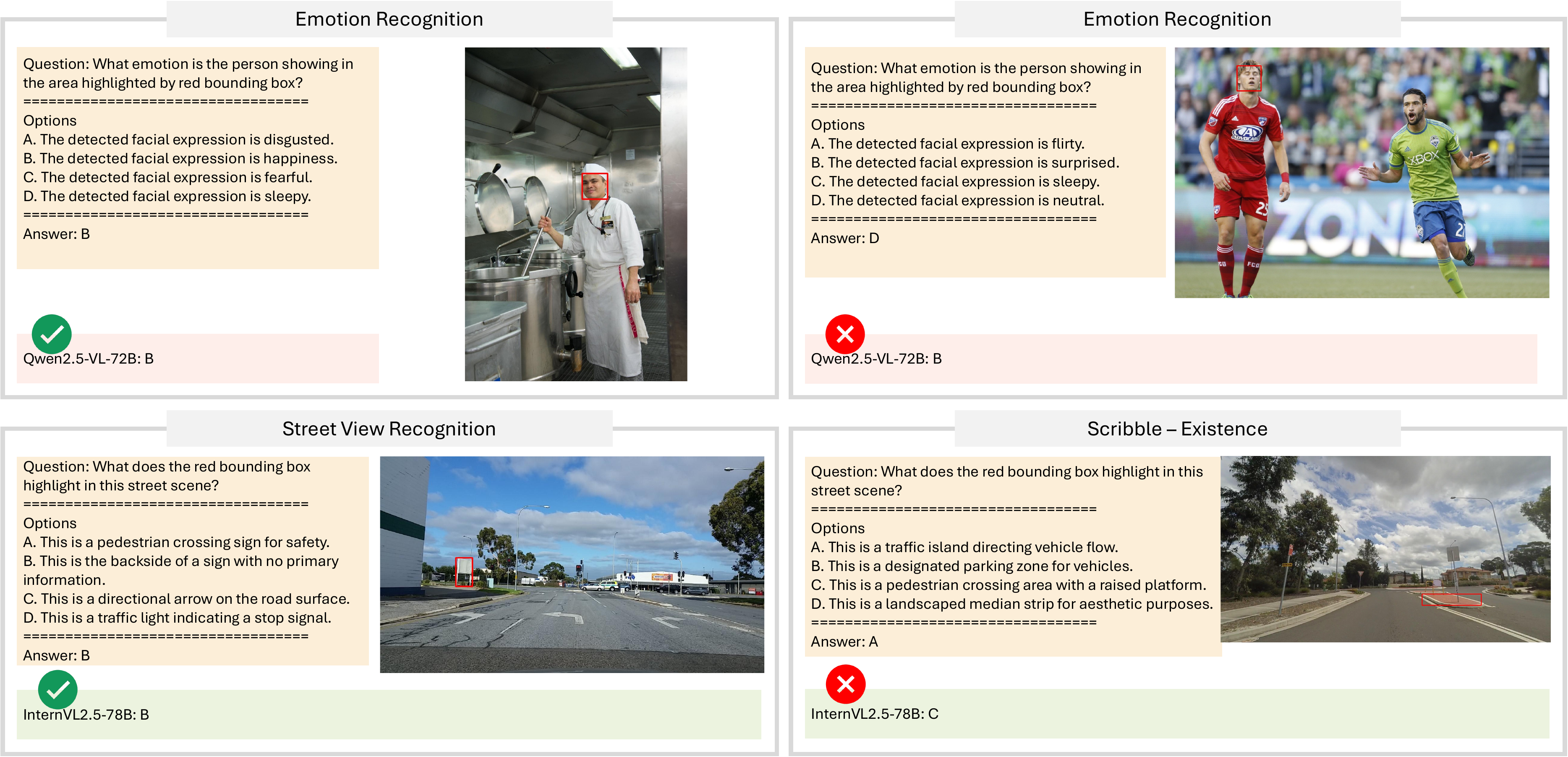}
    \caption{Qualitative results of facial emotion recognition task and street view recognition task.}
    \label{fig:stage-2-fer-svr}
\end{figure*}

\begin{figure*}[h]
    \centering
    \includegraphics[width=1\linewidth]{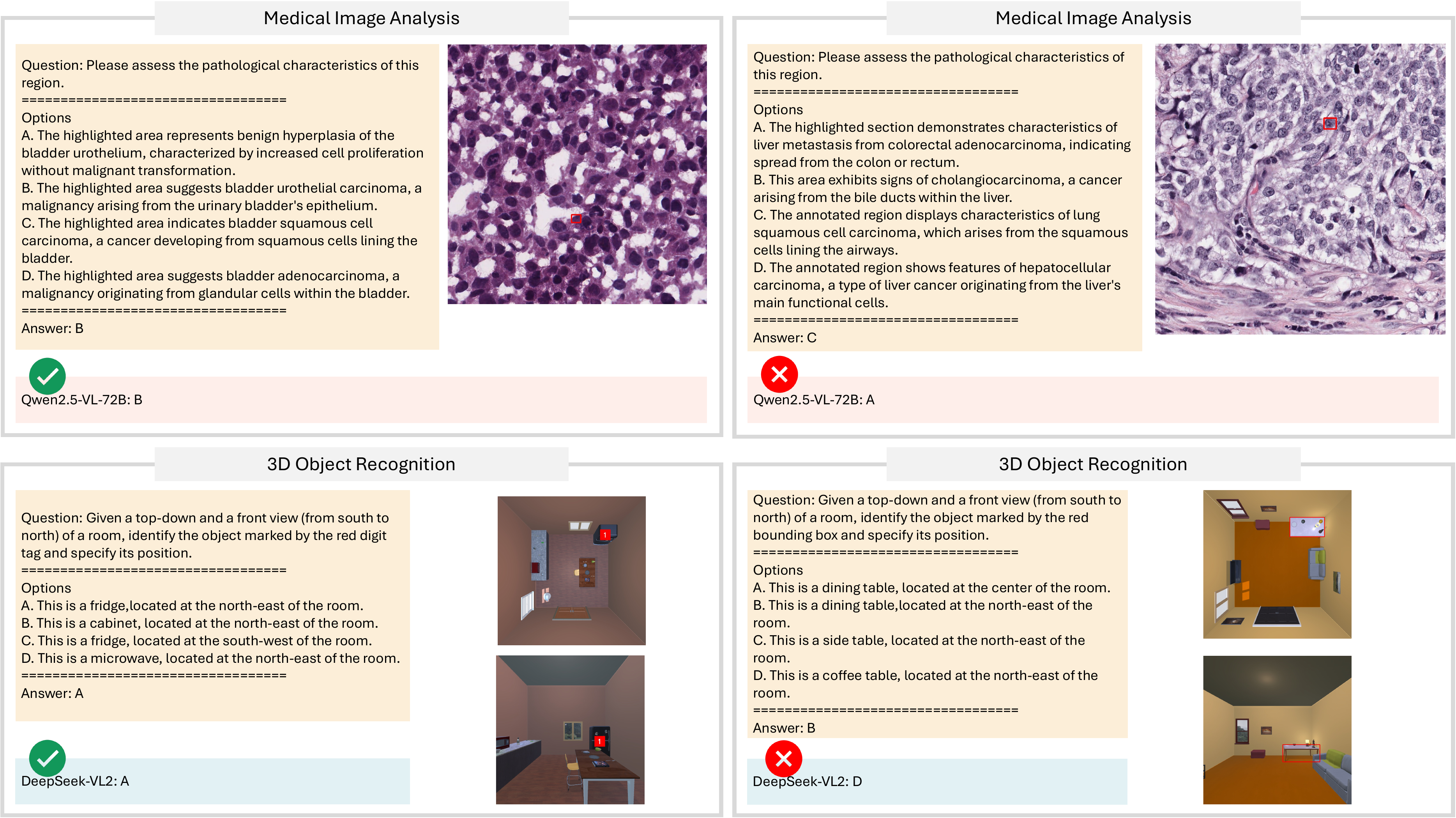}
    \caption{Qualitative results of medical image analysis task and 3D object recognition task.}
    \label{fig:stage-2-mia-3d}
\end{figure*}

\begin{figure*}[h]
    \centering
    \includegraphics[width=1\linewidth]{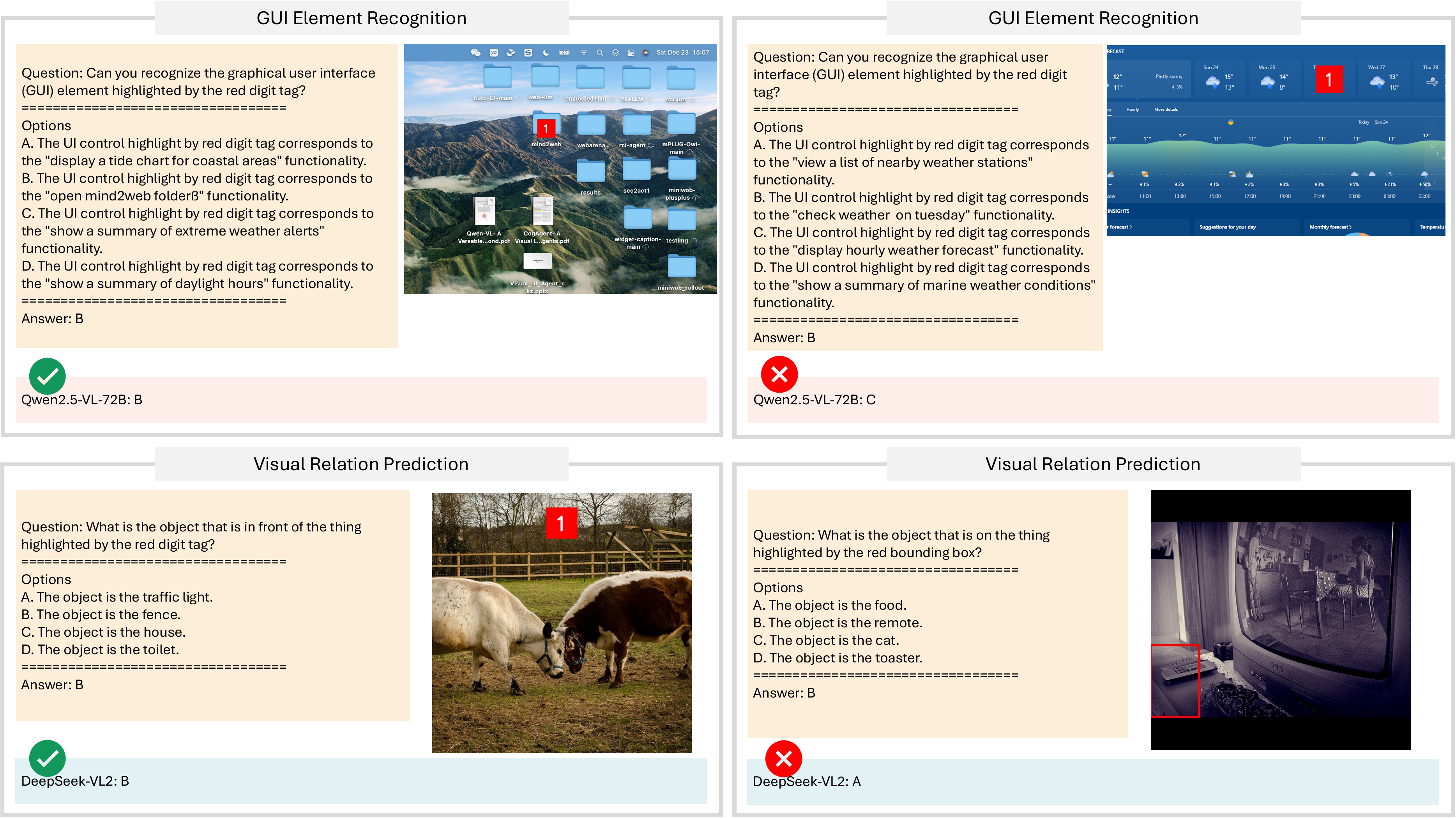}
    \caption{Qualitative results of GUI element recognition task and visual relation prediction object recognition task.}
    \label{fig:stage-2-gui-vrp}
\end{figure*}

In this section, we present qualitative visual results from our evaluation. In Stage 1, we assess the model's performance across all eight visual prompt types and five question types. Representative examples, illustrated in Figures \ref{fig:stage-1-arrow}, \ref{fig:stage-1-bbox}, \ref{fig:stage-1-contour}, \ref{fig:stage-1-tag}, \ref{fig:stage-1-mask}, \ref{fig:stage-1-oval}, \ref{fig:stage-1-scribble}, and \ref{fig:stage-1-point}, demonstrate how models interpret and respond to varied VPs. In Stage 2, we showcase qualitative results for downstream tasks, as shown in Figures \ref{fig:stage-2-fer-svr}, \ref{fig:stage-2-mia-3d}, and \ref{fig:stage-2-gui-vrp}, where model responses to task-specific questions are examined. These results provide key insights into the models' capabilities in processing multimodal inputs and highlight areas for further improvement.

\end{document}